\documentclass[10pt,twocolumn,letterpaper]{article}

\usepackage[pagenumbers]{cvpr} 

\usepackage{multirow}
\usepackage{tabularx}
\usepackage{scalerel}
\usepackage{lipsum}
\usepackage{pifont}
\usepackage{bm}

\definecolor{cvprblue}{rgb}{0.21,0.49,0.74}
\usepackage[pagebackref,breaklinks,colorlinks,allcolors=cvprblue]{hyperref}

\def\paperID{XXXX} 
\def\confName{CVPR}
\def\confYear{2026}

\title{What Matters for Scalable and Robust Learning \\ in End-to-End Driving Planners?}

\author{
David Holtz$^{1,2}$ \quad
Niklas Hanselmann$^{1}$ \quad
Simon Doll$^{1}$ \quad
Marius Cordts$^{1}$ \quad
Bernt Schiele$^{2}$ \\
[2mm]
$^1$~Mercedes-Benz AG\quad
$^2$~Max-Planck-Institute for Informatics, SIC\quad
}

\newcommand{\tf}[2]{{}^{#1}{\mathbf{T}_{#2}}}

\newcommand{\gray}[1]{\textcolor[rgb]{0.5, 0.5, 0.5}{#1}}
\newcommand{\myparagraph}[1]{\vspace{3pt}\noindent{\bf #1.}}
\newcommand{\mycaption}[2]{\caption{\textbf{#1.}\space #2}}

\begin{document}
\maketitle
\begin{abstract}
End-to-end autonomous driving has gained significant attention for its potential to learn robust behavior in interactive scenarios and scale with data.
Popular architectures often build on separate modules for perception and planning connected through latent representations, such as bird's eye view feature grids, to maintain end-to-end differentiability. 
This paradigm emerged mostly on open-loop datasets, with evaluation focusing not only on driving performance, but also intermediate perception tasks.
Unfortunately, architectural advances that excel in open-loop often fail to translate to scalable learning of robust closed-loop driving. 
In this paper, we systematically re-examine the impact of common architectural patterns on closed-loop performance:
(1) high-resolution perceptual representations,
(2) disentangled trajectory representations, and
(3) generative planning.
Crucially, our analysis evaluates the combined impact of these patterns, revealing both unexpected limitations as well as underexplored synergies.
Building on these insights, we introduce \textbf{BevAD}, a novel lightweight and highly scalable end-to-end driving architecture.
BevAD achieves 72.7\% success rate on the Bench2Drive benchmark and demonstrates strong data-scaling behavior using pure imitation learning.
Our code and models are publicly available here: {\footnotesize \url{https://dmholtz.github.io/bevad/}}

\end{abstract}
\section{Introduction}
\label{sec:intro}

\begin{figure}[htbp]
    \centering

    \begin{subfigure}{0.90\linewidth}
        \centering
        \includegraphics[width=\textwidth]{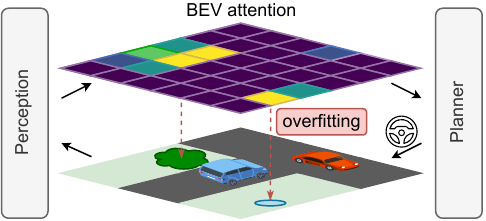}
        \caption{High-capacity perceptual representation, e.g., high-resolution BEV}
        \label{fig:sub1}
    \end{subfigure}
    \par 

    \begin{subfigure}{0.45\linewidth}
        \centering
        \includegraphics[width=\textwidth]{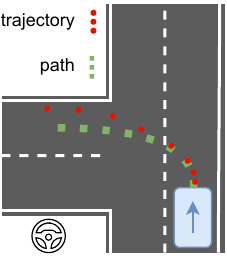}
        \caption{Planning Representation}
        \label{fig:sub2}
    \end{subfigure}
    \hfill 
    \begin{subfigure}{0.45\linewidth}
        \centering
        \includegraphics[width=\textwidth]{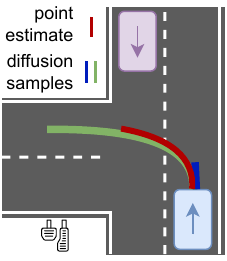}
        \caption{(Non-)generative Modeling}
        \label{fig:sub3}
    \end{subfigure}

    \mycaption{Architectural Patterns}{(a) High-resolution BEV features facilitate perception tasks, but promote overfitting the planner. (b) Closed-loop methods prefer path over trajectory representations due to robust steering. (c) Point-estimates interpolate between trajectory modes that diffusion-based sampling can breed.}
    \label{fig:main_figure}
\end{figure}

End-to-end autonomous driving (E2E-AD) has recently achieved great progress, driven by the possibility to optimize the entire stack in a planning-oriented manner~\cite{PlanningorientedAutonomous2023}. 
Compared to classical approaches with rule-based components, this enables human-like behavior in complex scenarios and promises performance gains that scale with data.
While E2E-AD exists in various flavors, popular approaches~\cite{PlanningorientedAutonomous2023, VADVectorizedScene2023, DualadDisentanglingDynamic2024, PARADriveParallelized2024} often implement modular but fully-differentiable transformer-based architectures with latent intermediate representations, such as bird's eye view (BEV) feature grids.
Popularized on open-loop benchmarks such as NuScenes~\cite{nuScenesMultimodalDataset2020}, these works typically do not evaluate in a closed-loop setting.
Unfortunately, approaches optimized for open-loop performance~\cite{PlanningorientedAutonomous2023, VADVectorizedScene2023, PARADriveParallelized2024} often fail to generalize in closed-loop driving scenarios~\cite{NAVSIMdatadriven2024, Bench2DriveTowardsMulti2024}, resulting in a divergence in the directions of architectural advances between works that solely focus on either setting.

In this paper, we take a step towards consolidating these advances in E2E-AD with a focus on closed-loop driving.
As depicted in Fig.~\ref{fig:main_figure} we systematically extend the design space proposed in ParaDrive~\cite{PARADriveParallelized2024}, by re-examining three architectural patterns: (1) The use of high-resolution perceptual representations as input to the planning module, more common to the open-loop setting~\cite{BEVFormerLearningBirds2022,PlanningorientedAutonomous2023}, (2) disentanglement of trajectories into lateral- and longitudinal components, mainly used in closed-loop driving~\cite{HiddenBiasesEnd2023,SimLingoVisionOnly2025} (3) the use of generative planners, previously underexplored for end-to-end closed-loop driving~\cite{DiffusionDriveTruncatedDiffusion2025, LookOutDiverseMulti2021, DiffusionBasedPlanning2025}. By evaluating these patterns jointly, we find that only one configuration admits robust scaling of performance.
In particular, we observe that high-resolution perceptual representations, shown to enable state-of-the-art (SotA) performance in open-loop~\cite{PlanningorientedAutonomous2023,VADVectorizedScene2023}, can be susceptible to causal confusion~\cite{Causalconfusionimitation2019}, and introduce a spatial bottleneck to mitigate this.
Furthermore, we show that disentangled trajectory representations and generative planning via diffusion~\cite{DiffusionBasedPlanning2025, DiffusionDriveTruncatedDiffusion2025}, previously studied only in isolation, provide complementary benefits in modeling multi-modal behavior, and see the strongest scaling properties when using both in conjunction.

Building on these insights, we develop BevAD.
It achieves SotA closed-loop driving performance on the challenging Bench2Drive benchmark~\cite{Bench2DriveTowardsMulti2024} based on the CARLA simulator~\cite{CARLAOpenUrban2017}, without any bells and whistles, and using camera sensors only.
Additionally, we demonstrate that BevAD strongly benefits from data scaling, with difficult skills emerging as the dataset size increases.
In summary, our core contributions are threefold, we:
\begin{enumerate}[(1)]
  \item Show that high-resolution perceptual representations can hinder learning robust planning and introduce a spatial bottleneck layer for mitigation.
  \item Analyze how a disentangled planning representation and diffusion-based planning provide complementary benefits for closed-loop driving.
  \item Integrate these insights and build BevAD, a lightweight and highly scalable E2E-AD architecture that achieves SotA closed-loop driving on Bench2Drive.
\end{enumerate}

\section{Related Work}
\label{sec:related-work}

\myparagraph{End-to-end Autonomous Driving}
Traditional autonomous driving (AD) stacks integrate standalone modules through compact predefined interfaces~\cite{functionalarchitectureautonomous2015}.
Although offering interpretability, this approach is prone to error accumulation, information loss between modules, and conflicting optimization objectives~\cite{PlanningorientedAutonomous2023}.
In contrast, the E2E-AD paradigm mitigates these limitations by optimizing the entire perception-to-planning pipeline, from raw sensor data to a planning trajectory or physical control commands~\cite{PlanningorientedAutonomous2023, EndEndLearning2016}.
UniAD~\cite{PlanningorientedAutonomous2023} established a foundational framework for \emph{modular} E2E-AD by introducing query-based mechanisms to jointly optimize perception, prediction, and planning.
A diverse landscape of stack-level designs has emerged, primarily categorized along three key dimensions:
\textit{(1) Task Selection.} This encompasses tasks such as 3D object detection~\cite{SparseDriveEndEnd2024}, multi-object tracking~\cite{PlanningorientedAutonomous2023, DualadDisentanglingDynamic2024}, online mapping~\cite{PlanningorientedAutonomous2023, VADVectorizedScene2023, SparseDriveEndEnd2024, PARADriveParallelized2024}, motion prediction~\cite{VADVectorizedScene2023, PlanningorientedAutonomous2023}, occupancy prediction~\cite{PlanningorientedAutonomous2023, SceneasOccupancy2023}, and increasingly, language-based tasks like commentary and visual question answering~\cite{SimLingoVisionOnly2025, ORIONHolisticEnd2025};
\textit{(2) Network Topology.} Information flow is governed by sequential~\cite{STP3End2022}, parallel~\cite{TransFuserImitationTransformer2023, PARADriveParallelized2024} and hybrid~\cite{PlanningorientedAutonomous2023, VADVectorizedScene2023, SparseDriveEndEnd2024} module placements or by unified transformers~\cite{HiPADHierarchical2025};
\textit{(3) Intermediate Representation.} These include sparse (e.g., instance-level)~\cite{SparseDriveEndEnd2024, DualadDisentanglingDynamic2024} and dense (e.g., BEV-centric~\cite{PARADriveParallelized2024} or occupancy-based~\cite{SceneasOccupancy2023}) representations.
Despite this extensive architectural landscape, our study uncovers a misconception about a subtle, yet pivotal detail within many existing designs:
High-resolution intermediate representations, such as an increased number of BEV features, have demonstrated improvements in open-loop perception~\cite{BEVFormerLearningBirds2022} and are posited to improve planning performance~\cite{Bench2DriveTowardsMulti2024}.
Conversely, we propose spatially compressing the perceptual representation before planner input and show substantially enhanced closed-loop driving robustness.
This finding corroborates observations that other successful closed-loop driving methods often employ architectural simplifications~\cite{SimLingoVisionOnly2025, HiddenBiasesEnd2024}, such as relying solely on single front-facing cameras.
We systematically shed light on the importance of the perceptual representation design and observe that 360-degree perception works if the intermediate representation is otherwise limited in bandwidth.

\myparagraph{Imitation Learning for Planning}
Learning to plan with an autonomous vehicle can be broadly categorized into two paradigms, (1) imitation learning (IL) via behavior cloning and (2) reinforcement learning.
Research on IL surged after pioneering work~\cite{EndEndLearning2016, EndEndDriving2018} and with the availability of large autonomous driving datasets~\cite{nuScenesMultimodalDataset2020, Towardslearningbased2024} and simulators for closed-loop testing, such as CARLA~\cite{CARLAOpenUrban2017}.
Conditioning the driving policy on a signal that represents the driver's intention in addition to environment observations has become a pivotal framework~\cite{EndEndDriving2018}, allowing control over the policy at test time with navigation commands or target points.
However, behavior cloning is prone to introducing undesired side-effects such as covariate shift~\cite{MitigatingCovariateShift2025} and causal confusion~\cite{Causalconfusionimitation2019}, which are hard to detect with open-loop based evaluation schemes, even with hardened metrics~\cite{NAVSIMdatadriven2024}.
Data augmentation~\cite{HiddenBiasesEnd2023, SimLingoVisionOnly2025} and world models~\cite{Modelbasedimitation2022, MitigatingCovariateShift2025} help mitigate these effects, though they do not solve them entirely.
Moreover, scaling up training data for IL has been shown to improve planning, following a power-law relationship in open-loop metrics, though these gains saturated in closed-loop driving~\cite{DataScalingLaws2025,DataScalingLaws2025a}.

Many current E2E-AD solutions operate as point estimators, directly regressing the final plan~\cite{TransFuserImitationTransformer2023, HiddenBiasesEnd2024, SparseDriveEndEnd2024} in waypoint or action spaces, or by selecting from predefined anchors~\cite{VADv2EndEnd2024}.
Both approaches can further benefit from trajectory post-processing~\cite{HydranextRobust2025, PlanningorientedAutonomous2023}.
An emerging alternative is generative models, particularly diffusion-based planners, which learn the distribution of future trajectories conditioned on the scene and a given command.
At test time, they sample from this conditional distribution.
Such diffusion-based planning can fit complex human driving distributions better than point estimators~\cite{DiffusionBasedPlanning2025,SceneDiffuserefficientcontrollable2024}, further enabling test-time guidance with driving style preferences~\cite{DiffusionBasedPlanning2025}.
Truncated schedules were proposed to mitigate the computational overhead from iterative denoising~\cite{DiffusionDriveTruncatedDiffusion2025}.
Despite demonstrating strong performance in open-loop benchmarks~\cite{DiffusionDriveTruncatedDiffusion2025}, diffusion-based planners have seen limited adoption in closed-loop methods.
Moreover, they have not been investigated in prior data scaling studies~\cite{DataScalingLaws2025, DataScalingLaws2025a}.
This work systematically compares point estimators and diffusion-based planners in closed-loop driving, revealing superior data scalability for diffusion-based approaches.

Open-loop E2E-AD methods often generate temporal waypoint trajectories~\cite{PlanningorientedAutonomous2023, VADVectorizedScene2023, DiffusionDriveTruncatedDiffusion2025, SparseDriveEndEnd2024, PARADriveParallelized2024}, thereby entangling lateral and longitudinal control.
This representation can lead to sparse and ambiguous supervision, particularly for dynamic, multi-modal intersection scenarios~\cite{HiddenBiasesEnd2023}.
In contrast, SotA closed-loop methods favor disentangled outputs: a future path independent of time and a target speed for longitudinal control~\cite{HiddenBiasesEnd2023, HiddenBiasesEnd2024, SimLingoVisionOnly2025, HiPADHierarchical2025}.
Both generative modeling and disentangled planning representations are patterns for learning multi-modal futures, yet their interaction remains underexplored.
Our study reveals strong synergies for scalable and robust end-to-end learning.

\section{Revisiting Common Architectural Patterns}

We briefly summarize common architectural patterns that were previously studied in isolation and occur predominantly either in the open-loop or the closed-loop setting.

\begin{enumerate}
    \item \textit{High-Resolution Perceptual Representations} are employed to improve performance in perception tasks~\cite{BEVFormerLearningBirds2022}, but are primarily studied in open-loop.
    While~\cite{Bench2DriveTowardsMulti2024} provides some evidence for benefits in closed-loop driving, leading closed-loop methods in CARLA inherently employ lower-capacity representations due to their reduced sensor configuration.
    We aim to systematically re-examine the impact of the BEV size on learning precise representations for closed-loop driving.
    \item \textit{Disentangled Planning Representations} are employed by the top-three closed-loop methods in CARLA as a measure to reduce the ambiguity of multi-modal futures.
    \item \textit{Generative Planners} emerged as a principled measure to model multi-modality in driving, but their study is mainly driven by open-loop methods.
\end{enumerate}

\noindent
In the remainder of this section, we first introduce our analysis framework (\cref{sec:analyis-framework}), along with the experiment setup (\cref{sec:experiment-setup}).
Subsequently, we analyze the impact of the perceptual representation (\cref{sec:high-capacity-interfaces}), the patterns for modeling multi-modal future jointly (\cref{sec:insufficient-modeling}) and the implications on scaling (\cref{sec:scalability}).

\begin{figure*}[htbp]
    \centering
    \begin{subfigure}[t]{0.44\textwidth}
        \centering
        \includegraphics[height=6.4cm]{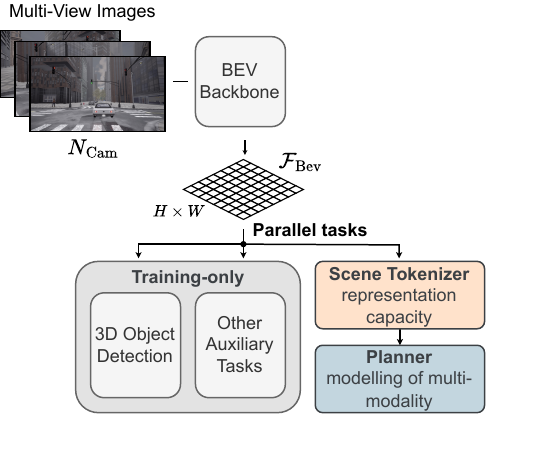} 
        \caption{Pipeline Overview}
        \label{fig:pipeline-overview}
    \end{subfigure}
    \hfill
    \begin{subfigure}[t]{0.55\textwidth}
        \centering
        \includegraphics[height=6.4cm]{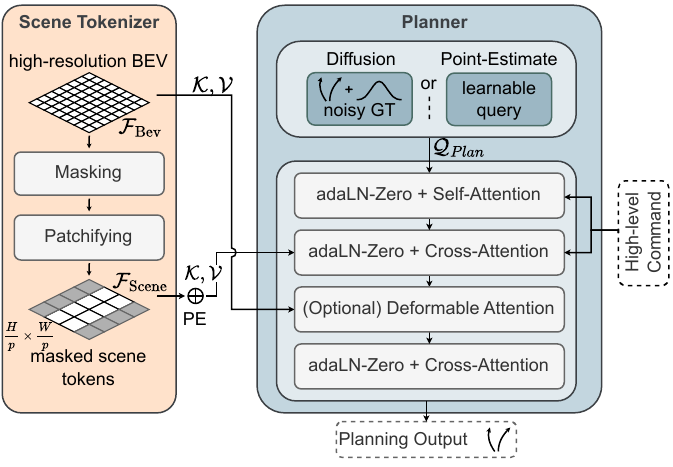} 
        \caption{Proposed Scene Tokenizer and Planning Head}
        \label{fig:planning-head}
    \end{subfigure}
    \mycaption{Analysis Framework}{(a) We build our analysis framework on ParaDrive~\cite{PARADriveParallelized2024}. (b) We introduce a scene tokenizer to reduce the spatial resolution of the BEV features. The design of our planning head is based on a diffusion transformer~\cite{ScalableDiffusionModels2023}. Crucially, the choice of the planning queries determines whether the planner is modeled as a point estimator or by diffusion.}
    \label{fig:analysis-framework}
\end{figure*}

\subsection{Analysis Framework}
\label{sec:analyis-framework}

Our analysis framework as shown in~\cref{fig:analysis-framework} is built upon ParaDrive~\cite{PARADriveParallelized2024} for the following key reasons:
(1) ParaDrive provides a systematic, module-level architecture for E2E-AD stacks, offering a well-defined foundation;
(2) Its planner operates independently from auxiliary task, facilitating a focused analysis of the perception-planner interface;
(3) ParaDrive's design, based on a real-world sensor configuration, has demonstrated strong performance on open-loop nuScenes tasks, unlike CARLA-specific methods.
Our framework prioritizes training efficiency to scale beyond nuScenes~\cite{nuScenesMultimodalDataset2020} to larger imitation learning datasets for CARLA~\cite{HiddenBiasesEnd2024, SimLingoVisionOnly2025}.
This is achieved through a streamlined pipeline (\cref{fig:pipeline-overview}), optimizing the BEV backbone and removing non-essential auxiliary tasks, while maintaining a realistic sensor setup.
Key aspects of the components are stated below, further details are found in the supplementary.

\myparagraph{BEV Backbone}
The BEV backbone processes images of $N_{\text{Cam}} = 6$ cameras to produce BEV features $\mathcal{F}_{\text{Bev}}$ with dimensions $H \times W$, comprising RADIO~\cite{RADIOv2.5ImprovedBaselines2025} with low-rank adapter~\cite{LoRALowRank2022} as its image backbone and a BEV encoder based on BEVFormer~\cite{BEVFormerLearningBirds2022}.
Significantly improved runtime is achieved by replacing the recurrent BEV feature generation with cached features streamed from short episode snippets during training~\cite{ExploringObjectCentric2023}.
Furthermore, we introduce a novel camera augmentation technique, applying a random transformation $\tf{\text{BEV}}{\text{car}}$ to all camera extrinsics to recover from compounding errors during closed-loop inference.

\myparagraph{Auxiliary Tasks}
We implement a DETR-style object decoder~\cite{EndEndObject2020} with deformable cross-attention~\cite{DeformableDETRDeformable2021} to supervise the BEV features during training~\cite{BEVFormerLearningBirds2022}.
We pruned other auxiliary tasks as used in~\cite{PlanningorientedAutonomous2023, PARADriveParallelized2024} as they did not demonstrably improve closed-loop performance during initial tests, but introduced significant runtime overhead.

\myparagraph{Planning Head}
We adopt a Transformer~\cite{Attentionisall2017} decoder architecture as depicted in \cref{fig:planning-head}:
Self-attention among planning queries $\mathcal{Q}_{\text{Plan}}$ enables mutual alignment, while cross-attention to scene tokens $\mathcal{F}_{\text{Scene}}$ allows extraction of global scene features, following~\cite{PlanningorientedAutonomous2023,PARADriveParallelized2024}.
Subsequently, we employ a coarse-to-fine strategy using an optional deformable attention layer, refining $\mathcal{Q}_{\text{Plan}}$ by sampling local, high-resolution BEV features $\mathcal{F}_{\text{Bev}}$.
Inspired by diffusion transformers~\cite{ScalableDiffusionModels2023}, each multi-head attention block is enclosed by adaLN-Zero transformations, incorporating conditioning from high-level driving commands, the ego-state, and optionally the diffusion timestep.
If the planner is a point estimator, the planning queries $\mathcal{Q}_{\text{Plan}}$ are implemented as learnable embeddings.
For diffusion-based planners, $\mathcal{Q}_{\text{Plan}}$ is generated by adding Gaussian noise to the ground truth according to a diffusion schedule such as DDIM~\cite{DenoisingDiffusionImplicit2021} and embedding the result into the transformer’s input space.
By reinterpreting the planning queries as path and velocity tokens instead of trajectory tokens and adjusting supervision accordingly, we can modify the planning representation.
This flexible design enables analysis across different formulations without altering the planner’s architecture.

\myparagraph{Controller}
Following~\cite{HiddenBiasesEnd2023, SimLingoVisionOnly2025, Bench2DriveTowardsMulti2024}, we employ two PID controllers to convert planning outputs into steering and acceleration commands.
The disentangled planning representation facilitates PID controller design by allowing separate processing of path and speed~\cite{HiddenBiasesEnd2023}.
To achieve the same for the trajectory representation, we fit a piecewise cubic Hermite polynomial to the temporal waypoints and interpolate at fixed distances, while speed is derived using a second-order difference quotient.
This allows consistent PID controller parameters across representations, minimizing the controller's critical impact on closed-loop driving.

\subsection{Experiment Setup}
\label{sec:experiment-setup}

\myparagraph{Data Collection}
There are currently two popular data sources for expert demonstrations in CARLA~\cite{CARLAOpenUrban2017} for training imitation learning models:
(1) Bench2Drive~\cite{Bench2DriveTowardsMulti2024} provides a dataset of expert demonstrations collected by the privileged, RL-based Think2Drive~\cite{Think2DriveEfficientReinforcement2024} expert along with sensor data and object annotations.
(2) The official CARLA leaderboard 2.0 benchmark provides specifications of long routes in CARLA with scenarios alongside, from which a dataset of expert demonstrations can be collected with the privileged, rule-based expert PDM-lite~\cite{DriveLMDrivingwithGraph2025, HiddenBiasesEnd2024}.
Simlingo~\cite{SimLingoVisionOnly2025} and TF++~\cite{HiddenBiasesEnd2024} split the long routes into shorter segments, each containing one scenario, and uniformly upsample routes with rare scenarios.
Due to various known label bugs in the Bench2Drive dataset, we adopt the second approach.
We re-collect training data for our six-camera sensor setup using the same route specifications as Simlingo and use these routes for training, unless stated otherwise.

\myparagraph{Training}
We conduct all experiments on 8xA100 80GB GPUs with a total batch size of $128$ in mixed-precision (bfloat16) to balance efficiency, memory usage and stability.
AdamW~\cite{DecoupledWeightDecay2017} Schedule-free~\cite{RoadLessScheduled2024} (learning rate: $2^{-4}$; weight decay: $0.01$) is used for optimization.
Our training consists of two stages:
A warm-up stage over four epochs to initialize the BEV backbone with perception supervision, followed by a second stage that adds planning supervision.
For faster convergence, we freeze the BEV backbone for all second-stage experiments except for studies on data scale.

\myparagraph{Benchmark and Metrics}
We perform closed-loop evaluations on the challenging Bench2Drive benchmark~\cite{Bench2DriveTowardsMulti2024} in CARLA~\cite{CARLAOpenUrban2017}.
Bench2Drive comprises 220 short test routes, each featuring a single scenario, enabling analysis of specific driving skills.
We report the official metrics driving score (DS) and success rate (SR).

\subsection{High-Resolution Perceptual Representations}
\label{sec:high-capacity-interfaces}

Established BEV-based end-to-end architectures connect perception and planning through $H \times W$ high-resolution latent BEV features~\cite{PlanningorientedAutonomous2023, PARADriveParallelized2024}.
We introduce a tokenizer (\cref{fig:planning-head}) that applies masking and patchifying to compress BEV features $\mathcal{F}_\text{Bev}$ into scene tokens $\mathcal{F}_{\text{Scene}}$, thereby channeling spatial information through a bottleneck.

\myparagraph{Masking}
We propose using a key padding mask in the global cross-attention of the planner to exclude BEV cells where planning queries $\mathcal{Q}_{\text{Plan}}$ cannot attend to.
Our initial experiments tested various masking strategies, such as removing distant parts to the left and right of the ego vehicle, and sophisticated masks based on the map segmentation outputs.
No significant differences were observed, so we use the simplest form, masking out 20\% of the left- and right-most BEV cells.
Although tailored to CARLA maps, this approach helps to determine if restricting the attention space facilitates learning a robust representation.

\myparagraph{Patchifying}
Inspired by Vision Transformers~\cite{ImageisWorth2021}, we propose pixel unshuffling for combining patches of $p \times p$ BEV features $\mathcal{F}_{\text{Bev}}$, (pixels) into spatial scene tokens $\mathcal{F}_{\text{Scene}}$, which our planner can globally attend to.
We prevent the channel dimension of the scene tokens from growing by $p^2$ by projecting the output of pixel unshuffling to a lower-dimensional space, thereby enforcing a bottleneck.
We explore $p \in \{1, 2, 4, 5\}$.
This analysis aims to understand how forced compression and sequence length in cross-attention impact learning a robust representation.

\begin{figure*}[htbp]
    \captionsetup[subfigure]{justification=centering}
    \centering
    \begin{subfigure}[b]{0.32\textwidth}
        \centering
        \includegraphics[width=\textwidth]{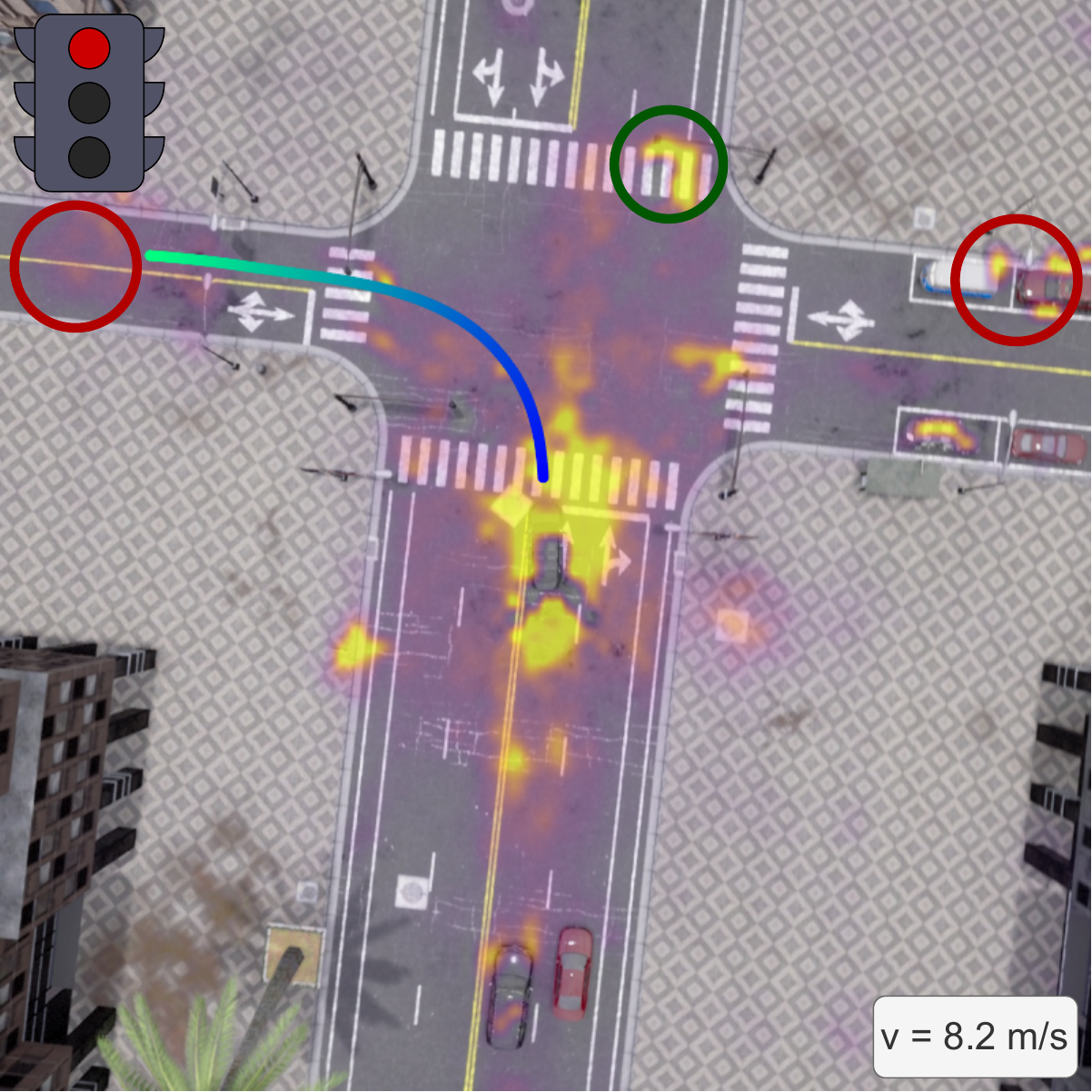}
        \caption{$p=1$, without masking}
        \label{fig:tok1-full}
    \end{subfigure}
    \hfill 
    \begin{subfigure}[b]{0.32\textwidth}
        \centering
        \includegraphics[width=\textwidth]{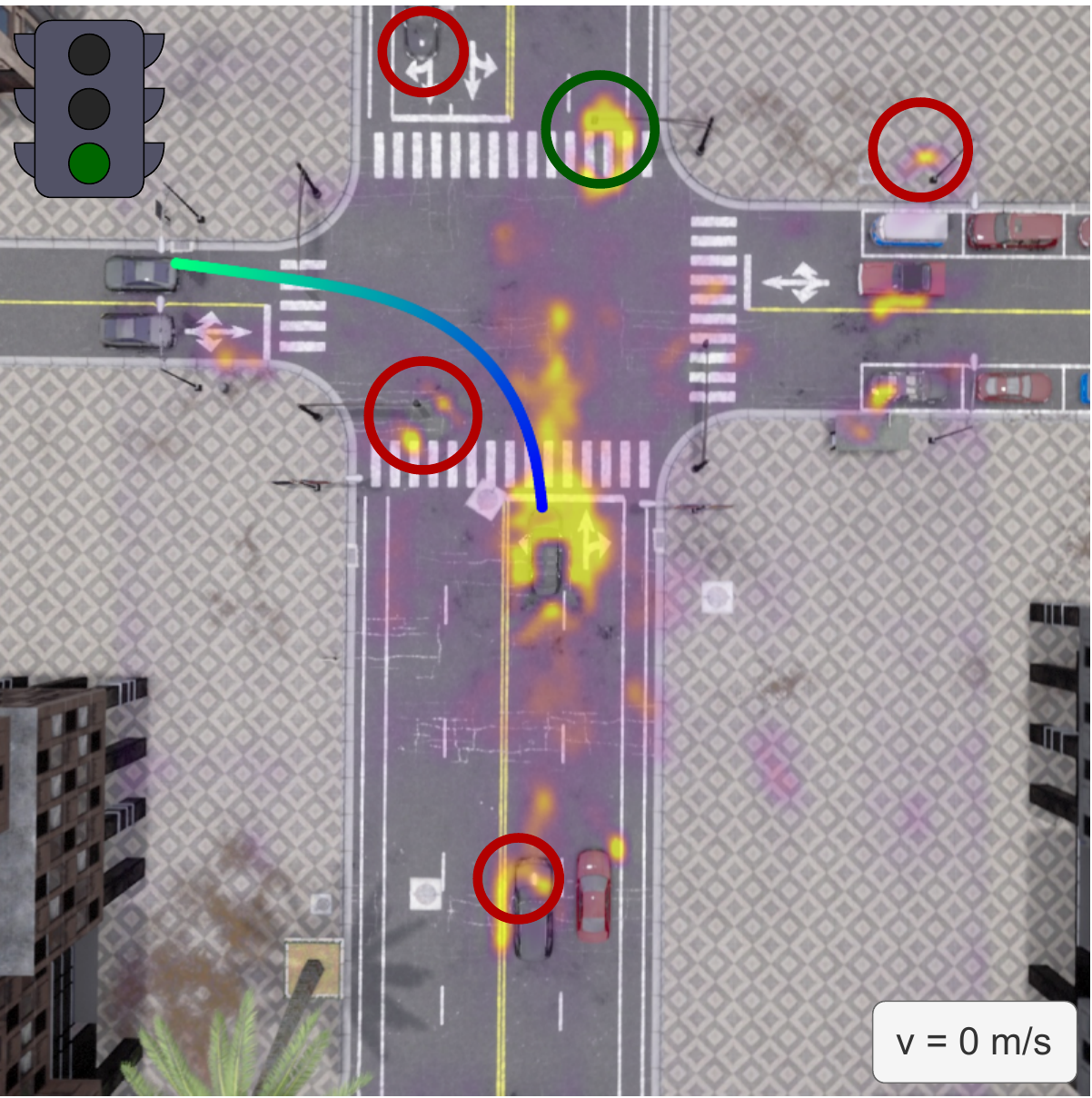}
        \caption{$p=1$, with masking}
        \label{fig:tok1-masked}
    \end{subfigure}
    \hfill 
    \begin{subfigure}[b]{0.32\textwidth}
        \centering
        \includegraphics[width=\textwidth]{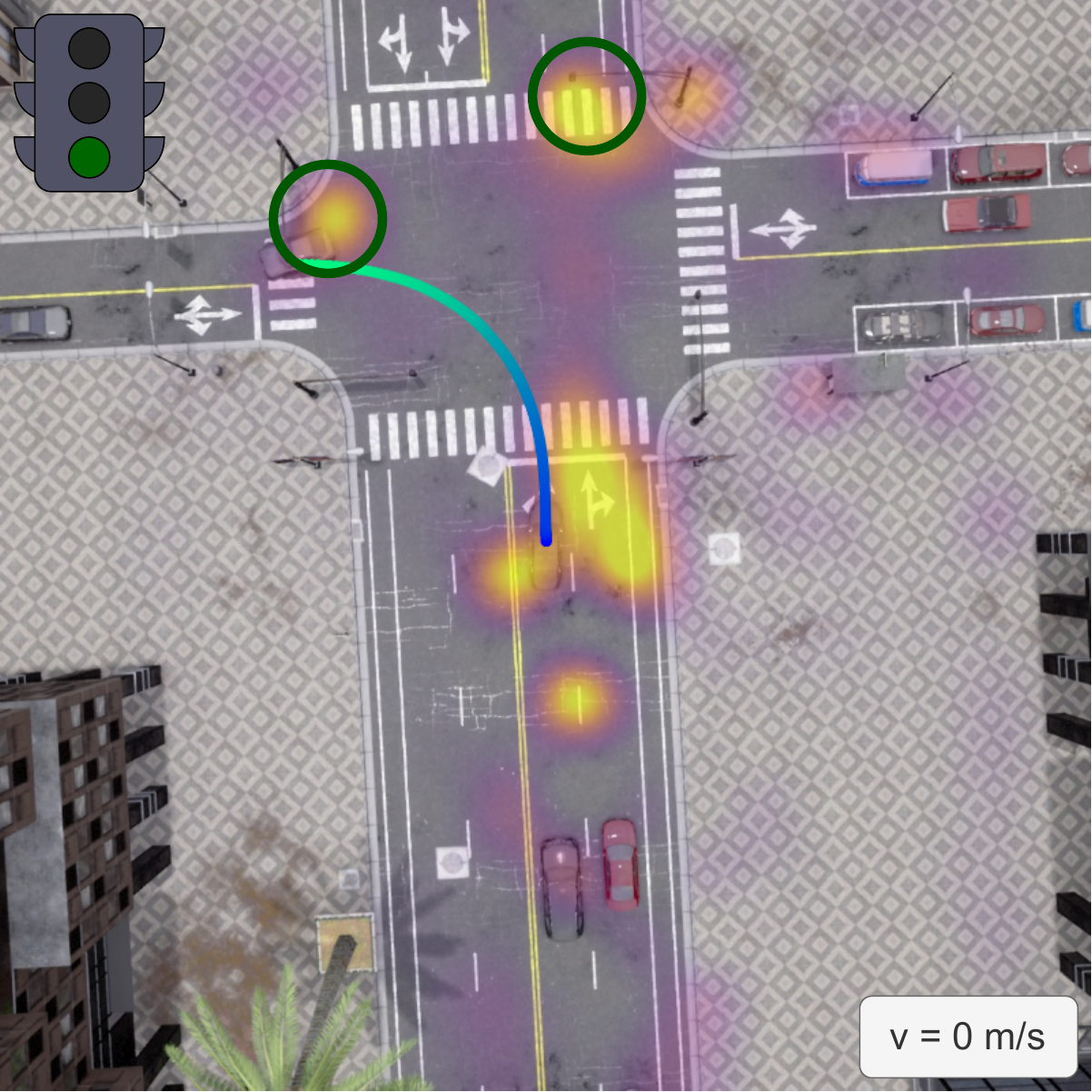}
        \caption{$p=4$, with masking}
        \label{fig:tok4-masked}
    \end{subfigure}
    \mycaption{Qualitative visualization of the planning queries' cross-attention to BEV features}{\cref{fig:tok1-full}. The planner attends to distant BEV cells. Despite strong attention on the traffic light, the autonomous vehicle runs the red light. \cref{fig:tok1-masked}: There are numerous attention spikes to random BEV cells, but barely no attention to the oncoming traffic. \cref{fig:tok4-masked}: The attention map significantly simplifies and exhibits fewer attention outliers.}
    \label{fig:attention}
\end{figure*}

\myparagraph{Results}
We employ a $100 \times 100$ BEV space, consistent with UniAD-tiny~\cite{PlanningorientedAutonomous2023, Bench2DriveTowardsMulti2024}, and a disentangled point-estimator planner~\cite{HiddenBiasesEnd2023}, aligning with SoTA on Bench2Drive~\cite{HiddenBiasesEnd2024, SimLingoVisionOnly2025}.
\cref{tab:bev-tokenizer} presents open- and closed-loop driving metrics for the tokenizer design space.
We observe significant improvements in closed-loop driving performance as the scene token count is reduced via masking and patchifying.
Specifically, restricting the planner's attention to masked BEV features enhances closed-loop driving, even when the mask is applied solely at test time.
Furthermore, summarizing $p \times p$ BEV feature patches into scene tokens reduces the planner's token count by a factor of $p^2$, yielding substantial closed-loop performance gains.
Despite the reduced BEV resolution, the L1 trajectory error marginally improves for $p \le 4$.
However, this compression strategy collapses for $p \ge 5$, resulting in a significant drop in both closed-loop and open-loop performance.

\myparagraph{Discussion}
Transformer-based models are known to struggle with identifying relevant information in long (text) sequences, even with modest token counts~\cite{BABILongTestingLimits2024}.
We relate this challenge to our setting, where high-resolution BEV inputs create long attention contexts.
We hypothesize that the planner overfits to spurious correlations in training data by deriving actions from memorized visual landmarks.
\cref{fig:attention} visualizes qualitative examples of planning query mean cross-attention activations.
In the absence of masking and patching, it reveals numerous punctual, high activation patterns in distant, often occluded or irrelevant BEV regions, strongly indicating causal confusion.
These learned shortcuts are not measurable by open-loop metrics like L1 due to averaging, but lead to catastrophic failures in distinct situations at test time.
By reducing the token count through masking and patchifying, our approach mitigates this causal confusion, significantly enhancing closed-loop driving by learning a more robust representation for test time.

Our finding contrasts with prior studies suggesting that higher BEV resolutions enhance downstream tasks such as 3D object detection~\cite{BEVFormerLearningBirds2022}.
This discrepancy stems from a fundamental difference between local detection and global planning tasks.
DeformableDETR-style detection heads leverage object locality by decoding queries to specific reference points~\cite{DeformableDETRDeformable2021, BEVFormerLearningBirds2022}.
While increased BEV resolution enhances localization precision, it does not expand a single query's receptive field.
In contrast, planning requires understanding critical scene elements that may not be localized near the immediate trajectory, thus necessitating global cross-attention~\cite{PlanningorientedAutonomous2023, PARADriveParallelized2024}.
In this global context, increasing BEV resolution expands the attention context size, contributing to the observed performance degradation.

\begin{table}
\setlength{\tabcolsep}{3.5pt}
\centering
\begin{tabular}{@{}ccc|ccc@{}}
\toprule
Mask                & Patch Size     & Scene Tokens    & DS $\uparrow$    & SR $\uparrow$ & L1 (m) $\downarrow$ \\ \midrule
\ding{55}           & 1              & $100 \times 100$& 66.86           & 36.36          & \gray{1.45}         \\
\ding{51}$^\dagger$ & 1              & $100 \times 60$ & 71.79            & 41.37         & \gray{1.45}         \\
\ding{51}           & 1              & $100 \times 60$ & 72.40            & 41.97         & \gray{1.53}         \\
\ding{51}           & 2              & $50 \times 30$  & 74.98            & 48.18         & \textbf{\gray{1.43}}\\
\ding{51}           & 4              & $25 \times 15$  & \textbf{82.62}   & \textbf{57.43}& \textbf{\gray{1.43}}\\
\ding{51}           & 5              & $20 \times 12$  & 66.44            & 40.91         & \gray{1.73}         \\ \bottomrule
\end{tabular}
\mycaption{Impact of Tokenizer Design}{\textit{Masking}: Restricting planning queries' attention benefits closed-loop driving. \textit{Patchifying}: Aggregating $p \times p$ BEV features into scene tokens significantly enhances closed-loop driving for $p \le 4$. Notably, these closed-loop improvements are not reflected in the open-loop L1 trajectory error metric. \textit{Legend:} $\dagger$ indicates test-time masking.}
\label{tab:bev-tokenizer}
\end{table}

\subsection{Modeling Multi-Modal Behavior}
\label{sec:insufficient-modeling}

The problem of inherent multi-modality in driving behavior is well-known in research~\cite{DiffusionBasedPlanning2025, HiddenBiasesEnd2023}.
Leading closed-loop methods in CARLA address this with a disentangled output representation that separates the spatial path from the speed profile instead of entangling them in a trajectory of temporal waypoints~\cite{HiddenBiasesEnd2024, SimLingoVisionOnly2025, HiPADHierarchical2025}.
Points on the path are obtained by sampling at fixed distances instead of fixed time intervals, which were shown to be less ambiguous, providing better supervision~\cite{HiddenBiasesEnd2023}.
Meanwhile, diffusion models~\cite{Denoisingdiffusionprobabilistic2020} can natively address the multi-modality in entangled temporal trajectories with generative modeling~\cite{DiffusionBasedPlanning2025, DiffusionDriveTruncatedDiffusion2025}.
On first glance, both patterns appear to solve a similar problem.
To discern the individual contributions and potential synergies of trajectory representation and (non)generative modeling, we systematically evaluate all four combinations.
As we observe that the DS and SR tend to obscure the distinct symptoms of driving failures, we additionally introduce static and dynamic infraction rates $\text{IR}_s$ and $\text{IR}_d$ for this experiment.
In a nutshell, $\text{IR}_s$ and $\text{IR}_d$ capture the prevalence of failures due to wrong path planning and inappropriate acceleration respectively; details can be found in the supplementary.
\begin{table}
\centering
\begin{tabular}{@{}cc|llll@{}}
\toprule
Model      & Repr. & DS $\uparrow$                   & SR $\uparrow$                   & $\text{IR}_{s}$ $\downarrow$ & $\text{IR}_{d}$ $\downarrow$ \\ \midrule
PE         & T     & 77.2 \small{$\pm 0.6$}          & 51.7 \small{$\pm 0.7$}          & 0.185                        & 0.505                        \\
PE         & P+S   & \textbf{82.6} \small{$\pm 1.0$} & \underline{57.4} \small{$\pm 0.6$}          & \textbf{0.055}               & 0.447                        \\
DI         & T     & 80.7 \small{$\pm 3.0$}          & 56.2 \small{$\pm 4.7$}          & 0.147                        & \textbf{0.391}               \\
DI         & P+S   & \underline{81.8} \small{$\pm 1.5$}          & \textbf{59.4} \small{$\pm 1.0$} & \underline{0.094}                        & \underline{0.423}                        \\ \bottomrule
\end{tabular}
\mycaption{Comparison of modeling and planning representation}{\textit{Legend:} PE: point-estimator, DI: diffusion, T: trajectory (entangled), P+S: path and speed (disentangled) representation.}
\label{tab:representations}
\end{table}

\myparagraph{Results}
As shown in~\cref{tab:representations} the disentangled representation significantly reduces static infractions, regardless of the modeling approach.
Particularly for point-estimators, this reflects strongly in the overall closed-loop scores, matching prior studies~\cite{HiddenBiasesEnd2023}.
We conclude that the disentangled representation is favorable for learning robust steering.
Generative modeling with diffusion reduces dynamic infractions, regardless of trajectory representation.
As a result, the entangled diffusion-based variant achieves similar overall SR than the disentangled point-estimator, though their failure modes are quite different.
Further, we observe complementary benefits for employing both diffusion-based modeling and disentangled representation, stated with the highest overall SR.
The lower driving score stems from -1.6\% route completion, since the diffusion model is less willing to make an infraction for the sake of route progress.

\subsection{Diminishing Returns when Scaling Non-Generative Planners}
\label{sec:scalability}
The promise of scaling performance with data is one of the main advantages of E2E-AD. Since generative planning can capture the full distribution of behavior, we hypothesize that it shows stronger benefits from scaling the dataset size. In the following, we hence examine the scaling behavior of diffusion- compared to point estimator-based planning.

\myparagraph{Scaling Data}
To scale training data beyond current datasets~\cite{SimLingoVisionOnly2025, Bench2DriveTowardsMulti2024}, we build a route generator that exhaustively plans short semantically plausible single scenario routes in all CARLA towns.
While being capable of building $>10^6$ unique route-scenario combinations (see supplementary), we only consider 8,000 uniformly sampled scenarios as additional training data in our scaling experiments.
For conducting data scaling experiments, we leverage established protocols from~\cite{DataScalingLaws2025, DataScalingLaws2025a}:
We create five training splits from the joint set of Simlingo's~\cite{SimLingoVisionOnly2025} and our routes, each approximately doubling in size.
These splits are cumulative (each being a subset of the larger ones~\cite{DataScalingLaws2025}), maintaining the same scenario distribution across all splits. We train all models on each data scaling split until convergence.

\begin{figure}
    \centering
    \includegraphics[width=0.9\linewidth]{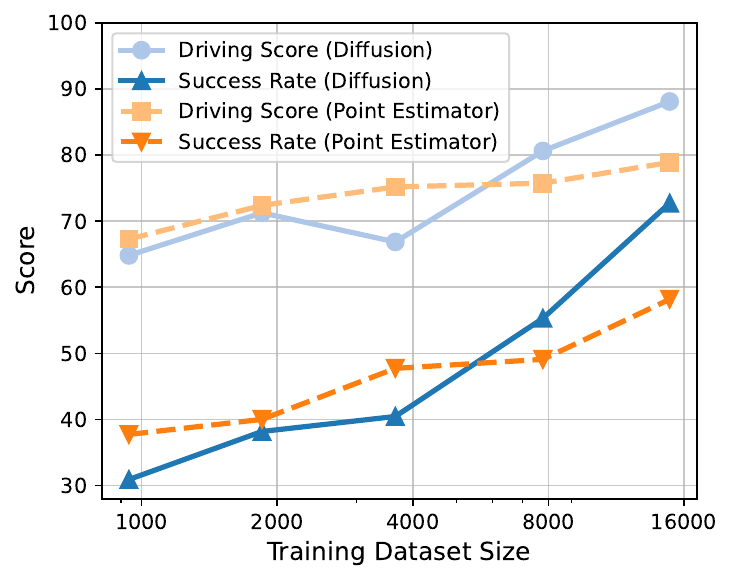}
    \mycaption{Scaling Properties}{Diffusion demonstrates superior performance over point estimators when scaled with sufficient training data, despite initially underperforming with limited data.}
    \label{fig:scaling-scores}
\end{figure}

\myparagraph{Results}
As reported in \cref{fig:scaling-scores}, both variants improve monotonically in SR as we double the training dataset.
In the low data regime, the point estimator slightly outperforms the diffusion-based planner.
After an inflection point (about 8000 training scenes), the growth rate decelerates, matching prior studies on closed-loop scaling laws for point estimate planners~\cite{DataScalingLaws2025a}.
On the other hand, diffusion-based planning maintains its linear rate of improvement until our largest data scaling split, and thereby substantially outperforms the point-estimator counterpart.
Interestingly, we cannot observe any saturation for the diffusion-based planner, unlike~\cite{DataScalingLaws2025, DataScalingLaws2025a} reported for closed-loop tests with point-estimator regressors.
This opens up opportunities for further improvements in the presence of larger datasets.

\myparagraph{Emerging Skills}
The scaling gains can also be broken down in terms of multi-ability evaluation protocol from Bench2Drive~\cite{Bench2DriveTowardsMulti2024}, where we observe that difficult skills emerge at larger training split sizes. For example, this is the case for the \emph{Give Way} and \emph{Merging} skills as required for the yielding scenario depicted in~\cref{fig:yield}. Detailed results can be found in the supplementary. 
\section{BevAD}

Integrating the above insights, BevAD emerges as a lightweight and highly scalable E2E-AD architecture from our analysis framework in \cref{fig:analysis-framework}.
It applies synergies of architectural patterns, previously studied in isolation, for dealing with multi-modality in driving and combats overfitting with effective BEV compression.
We compare BevAD to previous state-of-the-art in CARLA~\cite{CARLAOpenUrban2017}, providing a quantitative demonstration of BevAD's results (\cref{sec:sota}) along with qualitative results (\cref{sec:additional-results}) and real-world experiments on NAVSIM~\cite{NAVSIMdatadriven2024} (\cref{sec:real-world-results}).

\begin{table*}[t!]
\small
\centering
\begin{tabular}{@{}l|lll|cc@{}}
\toprule
\multirow{2}{*}{\textbf{Method}}                     & \multicolumn{3}{c|}{\textbf{Details}}             & \multicolumn{2}{c}{\textbf{Overall}}                                                \\ \cmidrule(l){2-6} 
                                                     & Expert      & Sensors       & Labels     & Driving Score $\uparrow$          & Success Rate $\uparrow$           \\ \midrule
VAD~\cite{VADVectorizedScene2023}                    & Think2Drive & 6x CAM        & O, M       & 42.35                             & 15.00                             \\
UniAD~\cite{PlanningorientedAutonomous2023}          & Think2Drive & 6x CAM        & O, M       & 45.81                             & 16.36                             \\ \midrule
ThinkTwice~\cite{ThinkTwiceDriving2023}              & Think2Drive & 6x CAM, LiDAR & O          & 62.44                             & 31.23                             \\
DriveAdapter~\cite{DriveAdapterBreakingCoupling2023} & Think2Drive & 6x CAM, LiDAR & O          & 64.22                             & 33.08                             \\
Hydra-NeXt~\cite{HydranextRobust2025}                & Think2Drive & 2x CAM        & -          & 73.86                             & 50.00                             \\
Orion~\cite{ORIONHolisticEnd2025}                    & Think2Drive & 6x CAM        & O, L       & 77.74                             & 54.62                             \\
TF++~\cite{HiddenBiasesEnd2024}                      & PDM-lite    & 1x CAM, LiDAR & O, M, S, D & 84.21                             & 67.27                             \\
Simlingo~\cite{SimLingoVisionOnly2025}               & PDM-lite    & 1x CAM        & L          & 85.07 \small{$\pm 0.95$}          & 67.27 \small{$\pm 2.11$}          \\
Hip-AD~\cite{HiPADHierarchical2025}                  & Think2Drive & 6x CAM        & O, M       & 86.77                             & 69.09                             \\
BridgeDrive$^\dagger$~\cite{BridgeDriveDiffusionBridge2025}    & PDM-lite    & 1x CAM, LiDAR & O, M, S, D & 86.87                             & 72.27                             \\ \midrule
BevAD-S \textit{(ours)}                              & PDM-lite    & 6x CAM        & O          & 80.63 \small{$\pm 1.76$}          & 55.30 \small{$\pm 2.63$}          \\
BevAD-M \textit{(ours)}                              & PDM-lite    & 6x CAM        & O          & \textbf{88.11} \small{$\pm 0.98$} & \textbf{72.73} \small{$\pm 1.98$} \\ \bottomrule
\end{tabular}
\caption{\textbf{Closed-loop Results on Bench2Drive.} Despite its simpler design BevAD outperforms previous modular baselines UniAD and VAD by a large margin, reaching SOTA-level performance. We highlight that BevAD can gain further substantial driving performance by uniformly scaling up training data. \textit{Legend:} O: 3D Object Detection, M: Map, S: Semantic Segmentation, D: Depth, L: Language, $\dagger$:~concurrent work. If available, we report mean and standard deviation over three seeds to account for the randomness in CARLA.}
\label{tab:b2d-sota}
\end{table*}

\subsection{Comparison to State of the Art}
\label{sec:sota}

A comprehensive comparison of BevAD to other methods on Bench2Drive with respect to training data, sensor configuration, supervision signals and performance can be found in \cref{tab:b2d-sota}.
For a fair comparison, our model is denoted as \mbox{BevAD-S} when trained solely on Simlingo routes~\cite{SimLingoVisionOnly2025}, and BevAD-M when trained with additional routes from our scaling study.
We consider UniAD~\cite{PlanningorientedAutonomous2023} and VAD~\cite{VADVectorizedScene2023} as baselines since their module-level architecture is most similar to BevAD.
We report the overall closed-loop driving score and success rate on the 220 test routes of Bench2Drive~\cite{Bench2DriveTowardsMulti2024}.
The significant improvements of +34.8 DS and +38.9 SR of BevAD-S compared to UniAD highlight the effectiveness of our tokenization as well as the complementary benefits of disentangled output representation and diffusion-based policy.
By uniformly scaling up training scenarios, BevAD-M outperforms all prior methods in terms of DS and SR, as well as the concurrent BridgeDrive~\cite{BridgeDriveDiffusionBridge2025} in terms of DS.
We refer to the supplementary for the more fine-grained multi-ability evaluation~\cite{Bench2DriveTowardsMulti2024} and analysis on driving skill evolution.

\subsection{Qualitative Results}
\label{sec:additional-results}
\begin{figure}[htbp]
    \centering
    \begin{subfigure}[t]{0.33\linewidth}
        \centering
        \includegraphics[height=5cm]{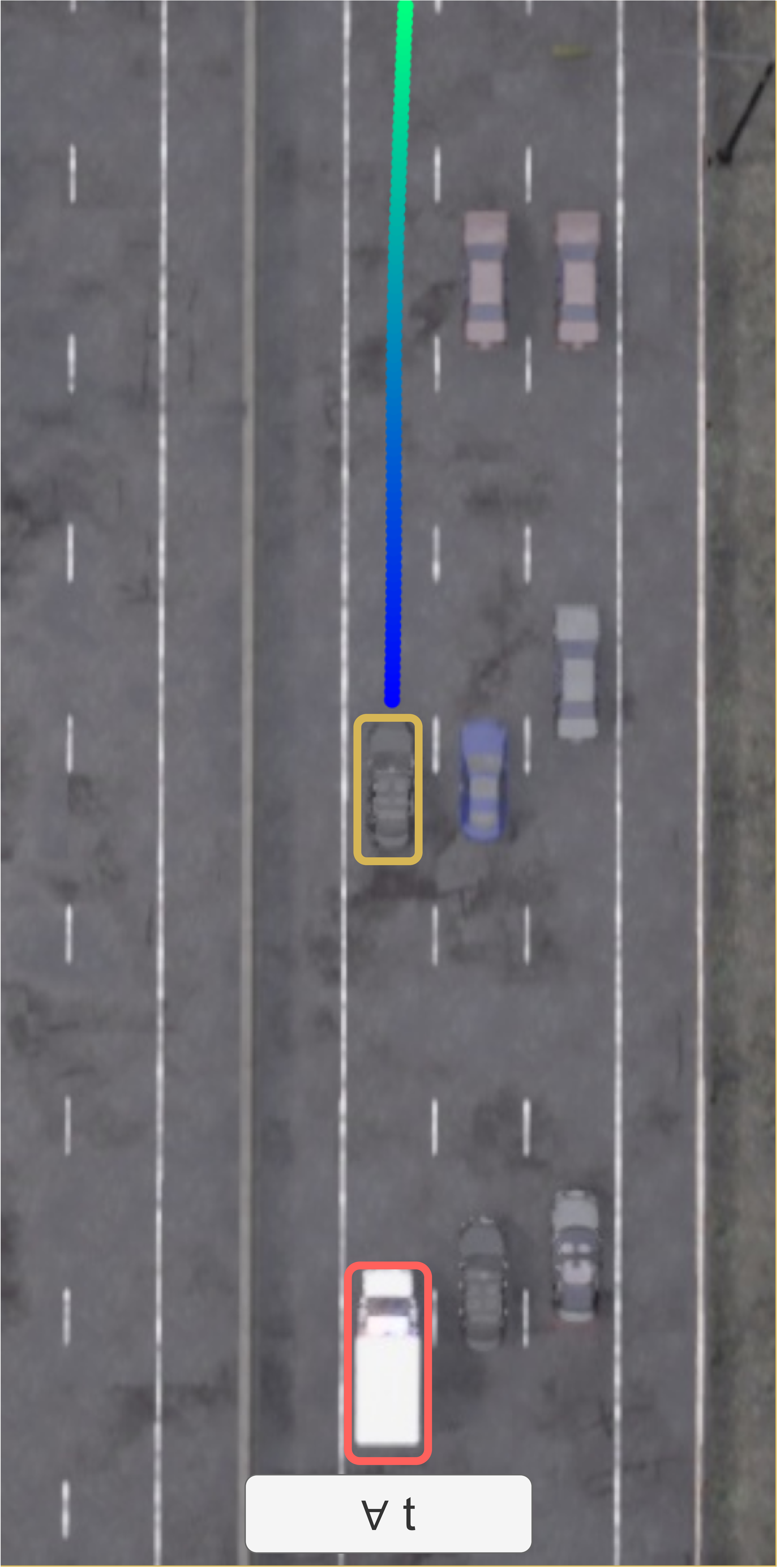}
        \caption{BevAD-S}
        \label{fig:subA}
    \end{subfigure}
    \hfill 
    \begin{subfigure}[t]{0.66\linewidth}
        \centering
        \includegraphics[height=5cm]{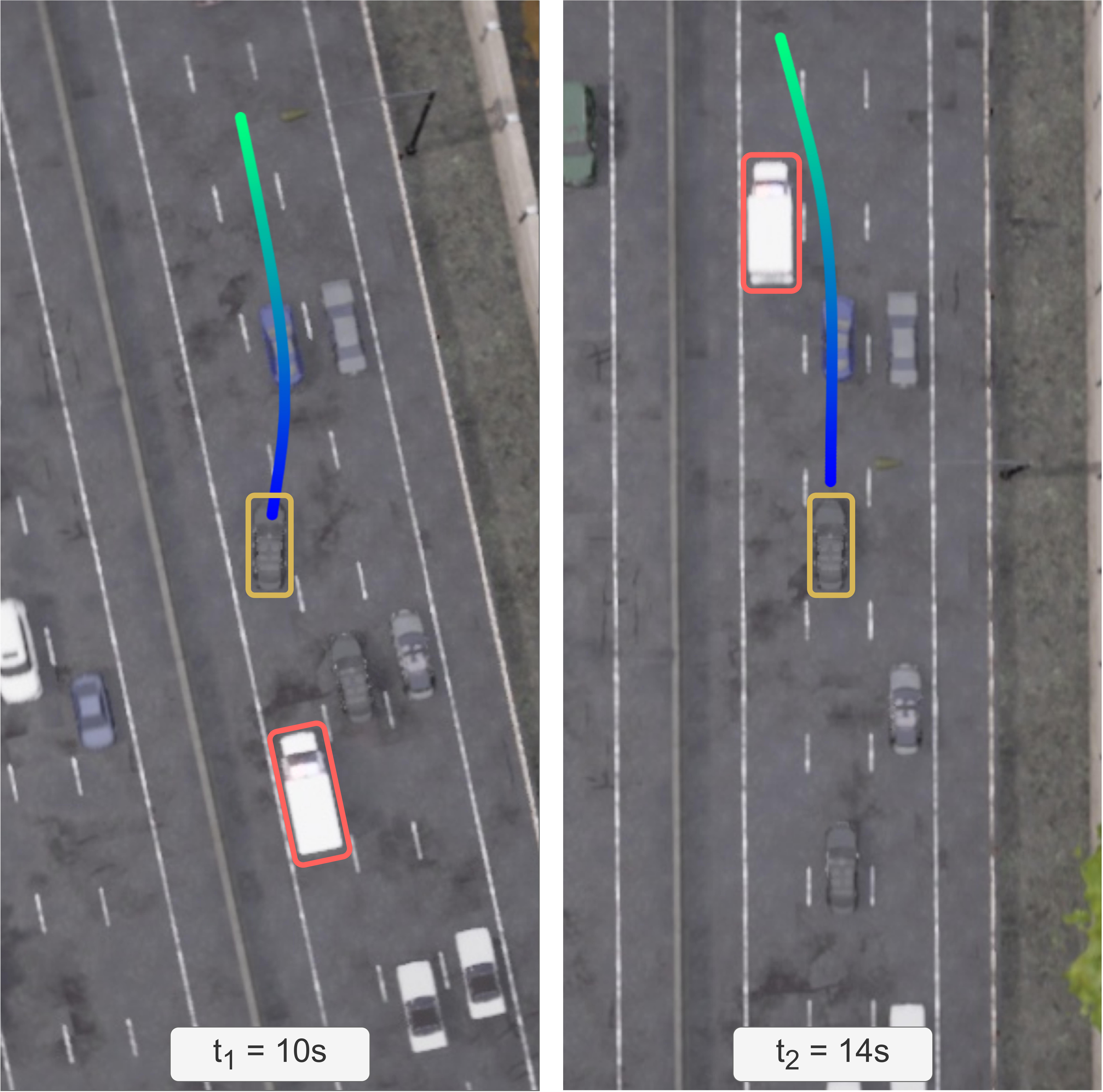}
        \caption{BevAD-M}
        \label{fig:subB}
    \end{subfigure}

    \mycaption{Yield to Emergency Vehicle}{By increasing the training dataset size, BevAD-M learns to yield to emergency vehicles (red) on highways by safely merging into slower traffic. This cability is absent at smaller data scales (BevAD-S) and in prior leading closed-loop methods\cite{SimLingoVisionOnly2025, HiPADHierarchical2025}.}
    \label{fig:yield}
\end{figure}

In the challenging \emph{YieldToEmergencyVehicle} scenario, BevAD demonstrates the ability to yield to a rapidly approaching emergency vehicle from behind.
\cref{fig:yield} illustrates that BevAD acquires this skill after scaling up training data.
Prior methods failed in such scenarios, either due to the lack of 360-degree camera perception~\cite{SimLingoVisionOnly2025} or insufficient training data~\cite{HiPADHierarchical2025}.
This underscores BevAD's effective utilization of its surrounding view BEV perception and its scalability.
Additional qualitative closed-loop demonstrations are provided in the supplementary material.

\myparagraph{Failure Cases}
We analyze common failure modes of BevAD-M:
\textit{(1)~Red Light Infractions} occur in 19\% of unsuccessful closed-loop runs.
For example, the driving model runs a red light in the PedestrianCrossing scenario after pedestrians have crossed, suggesting causal confusion.
\textit{(2)~Route Deviations} occur when BevAD ignores lane change commands, causing incorrect exits on multi-lane roads.
We attribute this to weak conditioning signals from navigation commands, which are often insufficient for timely lane changes. Strengthening conditioning with target points can mitigate this issue by guiding the model towards the correct lane center similar to~\cite{HiddenBiasesEnd2023}, though it increases reliance on precise map localization.
\textit{(3)~Miscellaneous Collisions} result from delayed reactions in time-critical scenarios or occur in situations that involve strong interaction with other vehicles, such as merging into flows.

\subsection{Real-world Experiments}
\label{sec:real-world-results}

We evaluate our method's real-world applicability on the NAVSIM planning benchmark~\cite{NAVSIMdatadriven2024}.
To match NAVSIM's expected planning representation, we adapt the diffusion planner to predict trajectories with associated yaw angles over a four-second horizon, and train BevAD end-to-end on the \texttt{navtrain} split for eight epochs.
We summarize performance on the \texttt{navtest} split in \cref{tab:navtest}, using the official NAVSIM metrics.
BevAD outperforms representative baselines UniAD~\cite{PlanningorientedAutonomous2023} and ParaDrive~\cite{PARADriveParallelized2024} by 3.2 and 2.6 PDMS, respectively, primarily due to improvements in drivable area compliance (DAC) and ego-progress (EP).
Notably, BevAD achieves this performance with only object detection and planning supervision, in contrast to baselines that also leverage online-mapping and occupancy prediction supervision.
BevAD's lightweight design yields a 570 GPU-hour (A100-80GB) training compute budget, 10x less than ParaDrive~\cite{NAVSIMdatadriven2024}.

Furthermore, we ablate the tokenizer design on real-world data:
As shown in \cref{tab:navtest}, removing masking degrades overall performance by 0.7 PDMS, while removing patchifying ($p=1$) results in a 1.0 PDMS degradation.
This demonstrates the effective generalization of our masking and tokenizing scheme to a real-world setting.

\begin{table}
\small
\setlength{\tabcolsep}{2pt}
\centering
\begin{tabular}{@{}l|cccc|c@{}}
\toprule
Method                                      & NC $\uparrow$ & DAC $\uparrow$    & TTC $\uparrow$    & EP $\uparrow$ & PDMS $\uparrow$ \\ \midrule
UniAD~\cite{PlanningorientedAutonomous2023} & 97.8          & 91.9              & 92.9              & 78.8          & 83.4 \\
ParaDrive~\cite{PARADriveParallelized2024}  & 97.9          & 92.4              & 93.0              & 79.3          & 84.0 \\ \midrule
BevAD \emph{(ours)}                         & \textbf{98.1} & \textbf{95.3}     & \textbf{94.5}     & \textbf{80.5} & \textbf{86.6} \\ \midrule
BevAD w/o mask                              & 97.8          & 94.9              & 93.7              & 80.3          & 85.9 \emph{(-0.7)} \\
BevAD w/o patchifying                       & 97.9          & 94.4              & 94.4              & 79.7          & 85.6 \emph{(-1.0)} \\ \bottomrule
\end{tabular}
\mycaption{Real-world Results on NAVSIM}{Performance comparison of BevAD against baseline methods on the real-world NAVSIM benchmark (\texttt{navtest}), including key ablations. \textit{Legend:} NC: no at-fault collision, DAC: drivable area compliance, TTC: time-to-collsion, EP: ego progress, PDMS: PDM score.}
\label{tab:navtest}
\end{table}

\section{Conclusion and Limitations}

We presented BevAD, a lightweight and highly scalable E2E-AD model that achieves SotA closed-loop driving on Bench2Drive.
BevAD emerges from our systematic analysis of common architectural patterns, previously studied in isolation.
We show that high-resolution BEV features can lead to overfitting, which we mitigate by forcing the planner to learn bottleneck.
Additionally, planning with diffusion complements disentangled planning output representations, particularly excelling when scaled with data.

We acknowledge several limitations.
First, while compressing the BEV along its spatial dimension significantly improved closed-loop driving, our approach may not directly extend to high-speed highway scenarios, which require long-range perception.
A principled, context-adaptive BEV masking strategy remains for future work.
Second, our analysis of failure cases suggests potential causal confusions.
Mitigating these, perhaps via incorporating world knowledge from VLMs or with reinforcement learning, requires further investigation.

\clearpage
\myparagraph{Acknowledgments}
This work is a result of the joint research project STADT:up (19A22006O).
The project is supported by the German Federal Ministry for Economic Affairs and Energy (BMWE), based on a decision of the German Bundestag.
The authors are solely responsible for the content of this publication.

{
    \small
    \bibliographystyle{ieeenat_fullname}
    \bibliography{main}
}

\clearpage
\setcounter{page}{1}
\maketitlesupplementary

\renewcommand{\thesubsection}{\Alph{subsection}}

\noindent
This supplementary material details our data scaling procedure (\cref{sec:suppl:dataset}), presents implementation details for our analysis framework and BevAD (\cref{sec:suppl:impl-details}), and includes additional quantitative and qualitative results (\cref{sec:suppl:results}).

\subsection{Dataset} \label{sec:suppl:dataset}

\subsubsection{Bench2Drive}

\begin{figure}[ht]
    \centering
    \includegraphics[width=1\linewidth]{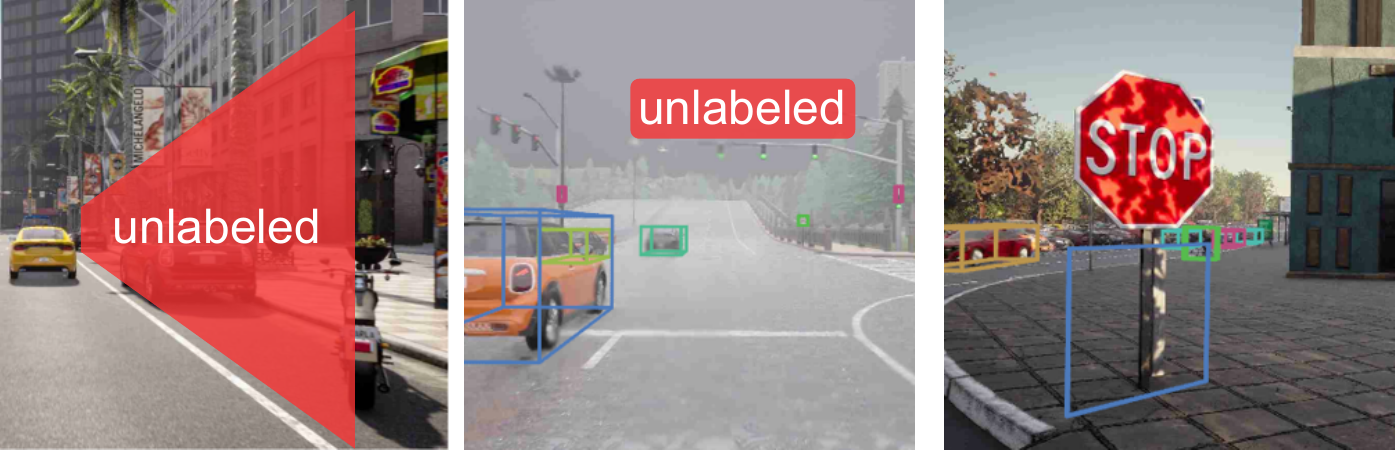}
    \mycaption{Label Flaws}{Bench2Drive exhibits incomplete and erroneous 3D bounding box annotations in various scenes.}
    \label{fig:label-bugs}
\end{figure}

Bench2Drive~\cite{Bench2DriveTowardsMulti2024} offers expert demonstrations from the Think2Drive expert~\cite{Think2DriveEfficientReinforcement2024}, comprising over 13,000 episodes for training.
However, the adoption of the dataset within the community is limited due to missing sensor modalities and language annotations~\cite{SimLingoVisionOnly2025, HiddenBiasesEnd2024, BridgeDriveDiffusionBridge2025}.
Furthermore, we identify significant 3D label flaws, as depicted in \cref{fig:label-bugs}: (1) Unlabeled parked vehicles in all CARLA towns except Town12 and Town13,
(2) Incomplete annotations for multi-lightbox traffic signals.
(3) Misplaced bounding boxes for numerous traffic signs and pedestrians.
These flaws create contradictory object detection supervision and enable planner shortcut learning, e.g., by distinguishing static cars from dynamic cars.
Critically, the lack of public route files or open-source expert code hinder dataset extension and issue resolution.
We therefore follow \cite{SimLingoVisionOnly2025, HiddenBiasesEnd2024, BridgeDriveDiffusionBridge2025} and utilize the CARLA~\cite{CARLAOpenUrban2017} Leaderboard 2.0 training routes and the rule-based PDM-lite expert~\cite{DriveLMDrivingwithGraph2025, HiddenBiasesEnd2024} for data collection.

\subsubsection{Route Generator}

Apart from the comprehensive Bench2Drive dataset, current CARLA imitation learning is limited by the diversity of expert demonstrations.
SotA methods \cite{HiddenBiasesEnd2024, SimLingoVisionOnly2025, BridgeDriveDiffusionBridge2025} utilize short, single-scenario route segments from CARLA Leaderboard 2.0 for training.
However, this approach suffers from significantly imbalanced scenario distributions (e.g., scenario InterurbanAdvancedActorFlow appears only six times, while others appear 40 times).
While these methods employ upsampling of rare routes and extensive camera- and weather augmentations, this mitigation is limited.
Fixed geographic contexts and similar actor behaviors can lead to overfitting to spurious correlations~\cite{PlanT2.0Exposing2025}.

To overcome these limitations, we build a novel route generator that creates unique route-scenario combinations within CARLA towns, enabling extensive and diverse data collection.
Our approach maintains the established concept of short, single-scenario training routes, which facilitates fine-grained control over scenario distribution; for simplicity, we adopt a uniform distribution.
The route generation algorithm proceeds in three steps: (1) Trigger Point Selection, (2) Route Planning and (3) Scenario Generation.

\myparagraph{Trigger Point Selection}
We employ a coarse-to-fine search strategy for trigger point selection.
Initially, scenarios are mapped to one of three coarse location classes: Intersection, No-Intersection, or Highway Ramp.
Subsequently, trigger points meeting these initial criteria are exhaustively sampled from all CARLA maps.
This candidate list is then refined using scenario-specific criteria.
For Intersections, relevant features include traffic lights, stop signs, turning options, bike lanes, and pedestrian crossings.
For No-Intersections, we consider the number of adjacent lanes (with/against ego traffic flow), and the presence of parking or shoulder lanes.
Highway Ramps are differentiated into On- and Off-Ramps.

\begin{figure}[htbp]
    \centering
    \begin{subfigure}[b]{0.4\linewidth}
        \centering
        \includegraphics[width=\textwidth]{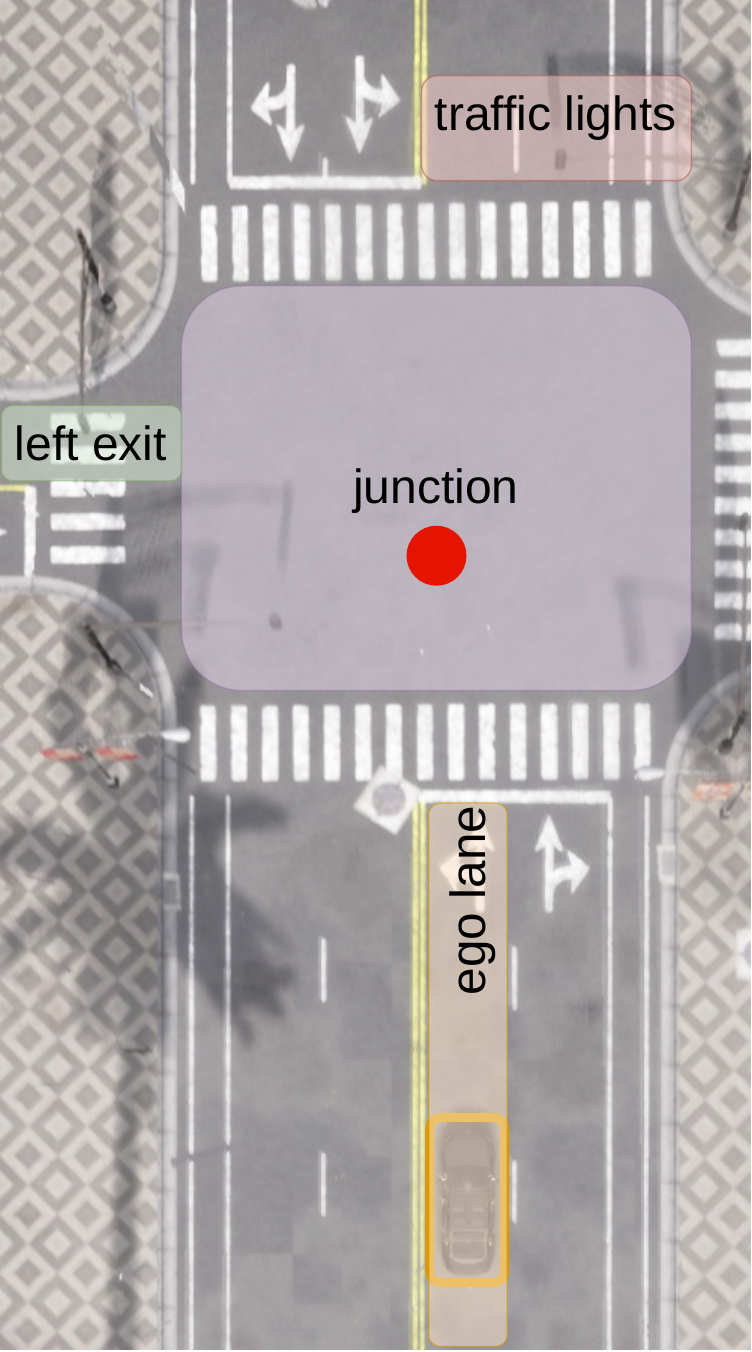} 
        \caption{Signalized Junction Left}
        \label{fig:subfigA}
    \end{subfigure}
    \hfill 
    \begin{subfigure}[b]{0.28\linewidth}
        \centering
        \includegraphics[width=\linewidth]{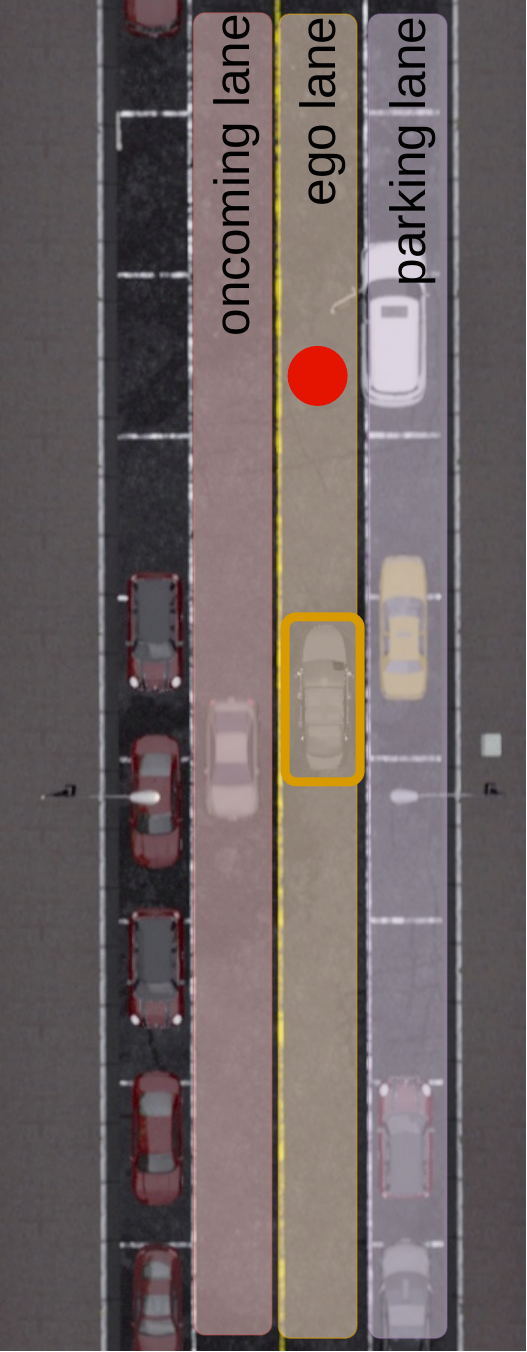} 
        \caption{Open Door}
        \label{fig:subfigB}
    \end{subfigure}
    \hfill 
    \begin{subfigure}[b]{0.28\linewidth}
        \centering
        \includegraphics[width=\linewidth]{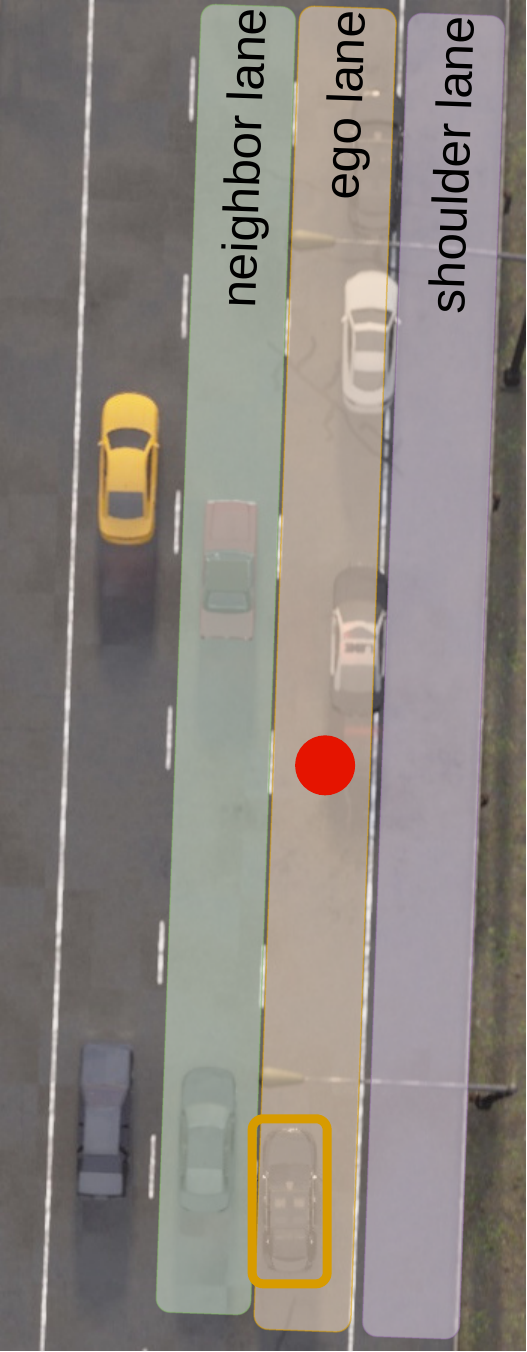} 
        \caption{Accident}
        \label{fig:subfigC}
    \end{subfigure}
    \mycaption{Visualization of Trigger Point Selection Criteria}{(a) This scenario requires a signalized junction with a left-turn exit relative to the ego lane. (b) The VehicleOpensDoorTwoWays scenario requires a right-side parking lane for the adversary and an adjacent left-side lane for oncoming traffic. (c) The Accident scenario demands a right-side shoulder lane for the group of blocking vehicles and an adjacent left-side lane with the same traffic flow.}
    \label{fig:tp-selection}
\end{figure}

For instance, the SignalizedJunctionLeftTurn scenario necessitates a signalized intersection with a left-turn option. The VehicleOpensDoorTwoWays scenario requires a right-side parking lane for the adversary and an adjacent lane with oncoming traffic.
Conversely, the Accident scenario demands a right-side shoulder lane and a left-side lane with traffic flowing in the same direction.
These examples are visualized in \cref{fig:tp-selection}.

\myparagraph{Route Planning}
We plan routes through a trigger point using a bidirectional search on the lane graph to determine start and end points.
This search is constrained to avoid additional intersections.
Distances between the start point, trigger point, and end point are randomly sampled from scenario-dependent intervals.
This strategy enhances variance and mitigates the learning of distance- and location-dependent shortcuts.

\myparagraph{Scenario Generation}
The final step involves configuring CARLA's pre-defined scenario behavior models at the trigger point.
To enhance diversity and prevent spurious correlations, additional scenario parameters are sampled from meaningful intervals, for instance, by varying inter-vehicle distances within traffic flows.

\pagebreak

Our route generator produces over 100,000 unique route-scenario combinations across all CARLA maps.
Due to the high cost of sensor data collection, we utilize a subset of 8,000 routes for our data scaling experiments.
Nevertheless, the generator's extensive diversity remains a significant asset for applications beyond imitation learning, such as reinforcement learning.

\subsection{Implementation Details} \label{sec:suppl:impl-details}

\subsubsection{Model}

\myparagraph{Efficient Streaming Training}
The training performance of previous BEV-based E2E-AD architectures is bottlenecked by recurrent BEV feature generation, required for fusing temporal information within the BEV encoder's temporal self-attention layer~\cite{BEVFormerLearningBirds2022}.
For instance, UniAD~\cite{PlanningorientedAutonomous2023} processes four past frames ($t-4, \dots, t-1$) at each training step $t$ with a frozen BEV backbone to eventually compute temporal self-attention between $\mathcal{F}_\text{Bev}^{(t-1)}$ and $\mathcal{F}_\text{Bev}^{(t)}$.
In contrast, we adapt streaming training from object-centric modeling~\cite{ExploringObjectCentric2023} to BEV-based architectures. Specifically, during training, we sample streams of $n$ subsequent frames, feeding them sequentially into the network.
A memory component caches current BEV features $\mathcal{F}_\text{Bev}^{(t)}$ at each step $t$, enabling subsequent steps ($t'=t+1$) to retrieve $\mathcal{F}_\text{Bev}^{(t'-1)} = \mathcal{F}_\text{Bev}^{(t)}$ from cache instead of recurrently recomputing it.
Short sequences (e.g., 2s, $n=20$ frames) are streamed to periodically reset the memory and mitigate non-orthogonal gradients~\cite{LearningStreamingVideo2025} arising from strongly correlated frames.
Our approach avoids four additional BEV backbone forward passes (compared to UniAD stage-1) and significantly reduces CPU load by requiring only one set of multi-view images per step, rather than five.

Additionally, we employ RADIO~\cite{RADIOv2.5ImprovedBaselines2025} as an image backbone with a low-rank adapter (LoRA)~\cite{LoRALowRank2022} for parameter-efficient finetuning.
RADIO provides rich, generic features, and LoRA finetuning significantly reduces VRAM requirements compared to backpropagating through a ResNet~\cite{DeepResidualLearning2016}.
Pruning additional heads for online-mapping, motion forecasting, and occupancy prediction further reduces BevAD's memory footprint.

These optimizations combined enable end-to-end training with a batch size of 16 on a single A100-80GB GPU, a significant improvement over UniAD, which required two-stage training with a frozen backbone due to VRAM constraints.
Our optimizations yield significantly higher training sample throughput, even surpassing UniAD-tiny, as shown in \cref{tab:training-efficiency}.
Specifically, BevAD processes $35 \times$ more training samples per second with a frozen BEV backbone, and achieves a $22 \times$ speed-up in end-to-end training.
These optimizations represent a major contribution towards scalable imitation learning for robust closed-loop performance, a benefit we extend to the community through our open-source code release.

\begin{table}
\centering
\begin{tabular}{@{}ll|c@{}}
\toprule
Method                        & BEV         & FPS (train) $\uparrow$ \\ \midrule
UniAD-tiny \cite{PlanningorientedAutonomous2023, Bench2DriveTowardsMulti2024}   & frozen  (stage-2) & 2.5   \\ \midrule   
\multirow{2}{*}{BevAD \textit{(ours)}}                                          & frozen            & 89.0  \\
                                                                                & end-to-end        & 55.0  \\ \bottomrule
\end{tabular}
\mycaption{Training Efficiency}{BevAD achieves substantial training speed-up relative to UniAD-tiny. Measured on 8xA100-80GB.}
\label{tab:training-efficiency}
\end{table}

\myparagraph{Camera Augmentation}
Camera augmentations are integral to robust closed-loop imitation learning~\cite{HiddenBiasesEnd2023}.
Perfect expert drivers, such as PDM-lite, maintain precise lane centering, leading to a training distribution dominated by ideal states.
However, during closed-loop testing, accumulated steering errors can cause the vehicle to drift, resulting in covariate shift~\cite{MitigatingCovariateShift2025} and degraded planner performance.
A common mitigation involves augmenting driver camera views with random shifts and rotations during training~\cite{HiddenBiasesEnd2023}, adopted by SotA methods on Bench2Drive~\cite{SimLingoVisionOnly2025, HiddenBiasesEnd2024, BridgeDriveDiffusionBridge2025}.
This augmentation simulates out-of-distribution states not observed with perfect expert driving.

However, these shift and rotation augmentations are typically limited to single front-facing camera setups and are challenging to apply to multi-camera systems.
Furthermore, their application to real-world data necessitates novel view synthesis, introducing pipeline overhead and potential artifacts.
To address these limitations, we propose a novel BEV-based augmentation strategy.
Instead of manipulating raw sensor data, we augment the BEV coordinate system directly.
This is achieved by sampling a random transformation matrix $\tf{Bev}{Car}$, comprising a small yaw rotation $\gamma \sim [-22.5^\circ, 22.5^\circ]$ and a lateral offset $\Delta y \sim [-0.75m, 0.75m]$.
This matrix is then applied to all camera transformation matrices $\tf{Car}{\text{Cam}_i}$:

\[\tf{\text{Bev}}{\text{Cam}_i} = \tf{\text{Bev}}{\text{Car}} \cdot \tf{Car}{\text{Cam}_i}\]

By providing $\tf{\text{Bev}}{\text{Cam}_i}$ to the BEV encoder, it builds BEV features $\mathcal{F}_\text{Bev}$ in the augmented BEV coordinate system rather than the vehicle coordinate system.
Additionally, we apply this transformation to all ground truth labels, i.e., 3D bounding boxes and planning trajectories.
Our augmentation prevents the detection head and planner from learning a bias towards axis-aligned objects or trajectories relative to the ego vehicle.
For example, the disentangled planner must learn to predict the future path in the augmented BEV space, necessitating a robust understanding of the BEV feature grid.
Unlike prior camera augmentations, our BEV-based scheme requires no augmented sensor data, as it only alters the generation of the latent BEV representation via a non-learnable, random transformation.

\myparagraph{Planning Queries}
The planning query's interpretation determines the output representation.
For an entangled trajectory representation, we define $\mathcal{P}_\text{Plan} = \mathcal{P}_\text{Traj} \in \mathbb{R}^{N_t \times N_\text{c}}$, where $N_t$ is the number of trajectory points and $N_\text{c}$ is the feature dimension.
For a disentangled representation, \mbox{$\mathcal{P}_\text{Plan} = [\mathcal{P}_\text{Path} \mathbin\Vert \mathcal{P}_\text{Speed}]$}, with $\mathcal{P}_\text{Path} \in \mathbb{R}^{N_p \times N_\text{c}}$ and \mbox{$\mathcal{P}_\text{Speed} \in \mathbb{R}^{N_t \times N_\text{c}}$}, where $N_p$ denotes the number of path points.
In our experiments, we set $N_t=15$ for a $3s$ planning horizon, resulting in temporal waypoint and speed predictions spaced at $0.2s$ intervals.
For path planning, we use $N_p=30$ with waypoints spaced $1m$ apart.

\begin{figure}
    \centering
    \includegraphics[width=0.7\linewidth]{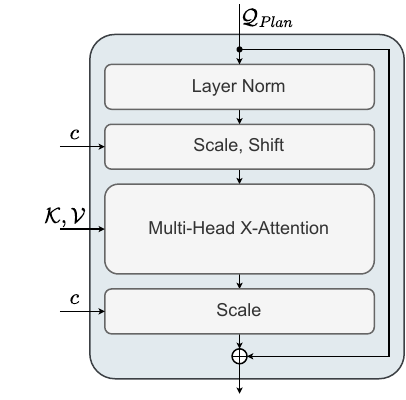}
    \mycaption{adaLN-Zero Attention}{We employ an adaLN-Zero transformation to condition the transformer on $\mathbf{c}$. For self-attention, we set \mbox{$\mathcal{K}=\mathcal{V}=\mathcal{Q}_\text{Plan}$} and for cross-attention, we set  $\mathcal{K}=\mathcal{V}=\mathcal{F}_\text{Scene}$.}
    \label{fig:adaln-attention}
\end{figure}

\myparagraph{Planning Head}
The planning head (\cref{fig:planning-head}) comprises $N_\text{layer}=8$ transformer decoder layers with a feature dimension of $N_\text{c}=512$.
Inspired by~\cite{ScalableDiffusionModels2023}, we integrate an adaLN-Zero transformation to condition the self-attention and global cross-attention layers with $\mathbf{c}$, as depicted in \cref{fig:adaln-attention}.
For the point estimator, \mbox{$\mathbf{c}=\text{emb}(\text{command}) + \text{emb}(v_0)$}, combining a learnable embedding of the high-level navigation command and a sinusoidal embedding of the current velocity.
For the diffusion-based planner, \mbox{$\mathbf{c}=\text{emb}(\text{command}) + \text{emb}(v_0) + \text{emb}(\tau)$}, additionally incorporating an embedding of the current diffusion timestep $\tau$.

\subsubsection{Diffusion-based Planning}

\myparagraph{Preliminaries}
Diffusion models are a class of generative models that learn to reverse a gradual noising process applied to data~\cite{Deepunsupervisedlearning2015, Denoisingdiffusionprobabilistic2020}.
They define a fixed forward Markov chain that progressively adds Gaussian noise to data, transforming it into a pure noise distribution.
Specifically, the forward process admits sampling of noisy data $\mathbf{x}_{\tau}$ for arbitrary (time)steps $\tau$ in the Markov chain in closed-form given clean data $\mathbf{x}_{0}$ by sampling Gaussian noise \mbox{$\bm{\epsilon} \sim \mathcal{N}(\mathbf{0}, \mathbf{1})$}~\cite{Denoisingdiffusionprobabilistic2020}:
\begin{align}
    \mathbf{x}_\tau(\mathbf{x}_0, \bm{\epsilon}) = \sqrt{\bar{\alpha_\tau}} \mathbf{x_0} + \sqrt{1-\bar{\alpha_\tau}}\bm{\epsilon} \label{eq:sampling-noise}
\end{align}
The sequence $\bar{\alpha_0}, \dots \bar{\alpha_T}$ typically stems from a variance schedule with constant parameters~\cite{Denoisingdiffusionprobabilistic2020, DenoisingDiffusionImplicit2021}.
The core of diffusion models lies in learning a reverse Markov chain that iteratively denoises samples over $T$ timesteps, starting from random noise \mbox{$p(\mathbf{x}_{T})=\mathcal{N}(\mathbf{x}_{T}; \mathbf{0}, \mathbf{I})$}, to recover data samples from the original distribution~\cite{Denoisingdiffusionprobabilistic2020}:
\begin{align}
    p_\theta(\mathbf{x}_{0:T}) := p(\mathbf{x}_{T})\prod_{\tau=1}^T p_\theta(\mathbf{x}_{(\tau-1)} \vert \mathbf{x}_\tau)
\end{align}
Instead of learning the Gaussian transitions $p_\theta(\mathbf{x}_{\tau-1} \vert \mathbf{x}_\tau)$ directly, it is common practice to learn a function approximator $f_\theta:(\mathbf{x}_\tau, \tau)\rightarrow \mathbf{x}_0$ parameterized by a neural network with learnable weights $\theta$~\cite{UnderstandingDiffusionModels2022}.
This allows learning the diffusion model by minimizing \cref{eq:diffusion-loss} for all timesteps $\tau$~\cite{Denoisingdiffusionprobabilistic2020}:
\begin{align}
   \left\| \mathbf{x}_0 - f_\theta(\mathbf{x}_\tau, \tau)\right\| \label{eq:diffusion-loss}
\end{align}

\noindent
For diffusion-based planning, the clean planning ground truth $\mathbf{x}_0$ is defined based on the planning representation:
\begin{itemize}
    \item Entangled: a series of normalized trajectory waypoints, $\mathbf{x}_0 = \{(x_i, y_i\}_{i=1}^{N_t}\}$.
    \item Disentangled: a tuple comprising normalized path waypoints and a normalized speed sequence, i.e., \mbox{$\mathbf{x}_0 = (\{(x_i, y_i\}_{i=1}^{N_p}, \{(v_i\}_{i=1}^{N_t}\})$}.
\end{itemize}

\myparagraph{Training}
During training, we sample a timestep $\tau$ from a uniform distribution and Gaussian noise $\bm{\epsilon} \sim \mathcal{N}(\mathbf{0}, \mathbf{1})$.
We obtain the noisy sample  $\mathbf{x}_\tau$ using \cref{eq:sampling-noise} and following the DDIM variance schedule~\cite{DenoisingDiffusionImplicit2021}.
The planning head serves as the \emph{conditional} function approximator $f_\theta:(\mathbf{\tilde{x}}_\tau, \tau,  \mathbf{z})$ for the reverse process~\cite{DiffusionDriveTruncatedDiffusion2025}, tasked with predicting $\mathbf{x}_0$ from $\mathbf{\tilde{x}}_\tau$, the diffusion timestep $\tau$, and conditioning context \mbox{$\mathbf{z}=(\mathcal{F}_\text{Scene}, c)$}.
Note that $\mathbf{x}_\tau$ is embedded into a latent space via an MLP to obtain $\mathbf{\tilde{x}}_\tau$ before being fed to $f_\theta$.

\myparagraph{Inference}
For inference, we start with a random sample $\mathbf{x}_T \sim \mathcal{N}(0,1)$ and iteratively denoise it using the trained function approximator $f_\theta(\mathbf{\tilde{x}}_\tau, \tau, \mathbf{z})$ and the DDIM sampling algorithm~\cite{DenoisingDiffusionImplicit2021}.
This process progressively denoises $\mathbf{x}_T$ to yield the final clean prediction $\mathbf{x}_0$.
We specifically employ DDIM's accelerated generation that conducts denoising with a subset of $S<T$ denoising steps to enhance computational efficiency during sampling.

\myparagraph{DiffusionDrive}
We do not adopt DiffusionDrive's~\cite{DiffusionDriveTruncatedDiffusion2025} truncated diffusion schedule in our experiments.
This decision stems from our observation, that DiffusionDrive's forward process adds noise to a fixed set of anchor trajectories, while the reverse process aims to predict ground truth trajectories, leading to an asymmetric reversal.
This issue is also thoroughly discussed in concurrent work~\cite{BridgeDriveDiffusionBridge2025}, which proposes a theoretically sound diffusion bridge formulation as a solution.

\subsubsection{Controller}
We adopt the disentangled PID controllers from Simlingo~\cite{SimLingoVisionOnly2025} for lateral and longitudinal control.
All controller parameters are retained, with the exception of the steering proportional gain, which is slightly lowered to $P_\text{steer}=1.8$ to reduce oversteering in our closed-loop agent.

\subsubsection{Loss}
The overall loss function for end-to-end training combines terms for 3D object detection and planning:
\begin{align}
\mathcal{L} = \lambda_{\text{det}} \mathcal{L}_\text{det} + \lambda_{\text{plan}} \mathcal{L}_\text{plan}
\end{align}
The detection loss $\mathcal{L}_\text{det}$, adopted from \cite{BEVFormerLearningBirds2022}, includes classification and regression components.
The planning loss $\mathcal{L}_\text{plan}$ varies with the planner's modeling choice:
\begin{itemize}
    \item For regression-based planning with a point estimator, $\mathcal{L}_\text{plan}$ is a smooth $L^1$ loss applied to the trajectory error, or path and speed error.
    \item For diffusion-based planning, $\mathcal{L}_\text{plan}$ is a smooth $L^1$ loss on the $\mathbf{x}_0$-prediction error, as defined in \cref{eq:diffusion-loss}.
\end{itemize}
The loss coefficients are set to $\lambda_\text{det}=1$ and $\lambda_\text{plan}=100$ to balance the magnitudes of the detection and planning loss components.

\subsubsection{Metrics}
To better characterize closed-loop failure modes, we introduce auxiliary metrics: the static infraction rate ($\text{IR}_\text{s}$) and dynamic infraction rate ($\text{IR}_\text{d}$).
\begin{align}
\text{IR}_\text{s} = \frac{N_\text{layout-collision}+N_\text{outside-lane}}{N_\text{routes}}
\end{align}
This metric quantifies infractions related to lateral control errors, such as collisions with static layout elements or driving outside the lane.
\begin{align}
\text{IR}_\text{d} = \frac{N_\text{actor-collision}+N_\text{red-light}+N_\text{stop-sign}}{N_\text{routes}}   
\end{align}
This metric captures infractions arising from longitudinal control errors and interactions with dynamic elements, including collisions with actors, running red lights, or failing to stop at stop signs.
Collectively, these metrics provide the expected number of infractions per route, offering a granular understanding of control deficiencies.

\subsection{Results} \label{sec:suppl:results}
This section presents an ablation study, quantitative results on the Bench2Drive benchmark, and qualitative demonstrations of BevAD's closed-loop driving.

\subsubsection{Ablations}
Our ablation studies investigate early design choices.
For the following ablations of camera augmentation and BEV size, our baseline is a planning head with a tokenizer ($p=4$, masking), disentangled representation, and a point-estimator regressor, as detailed in \cref{sec:high-capacity-interfaces} and \cref{tab:bev-tokenizer}.
Furthermore, we ablate the number of denoising steps in the diffusion planner using our strongest model, BevAD-M.

\myparagraph{Camera Augmentation}
To assess the effectiveness of our novel BEV-based augmentation, we compare the baseline against a variant trained without it.
As shown in \cref{tab:ablation-augmentation}, the absence of camera augmentation significantly degrades closed-loop performance.
This degradation primarily stems from a $5.7 \times$ increase in static infractions, leading to reduced route completion and increased secondary collisions.
These results underscore the critical role of our augmentation scheme in promoting robust driving and enabling recovery from compounding steering errors.
The results also highlight the insufficiency of common open-loop metrics, as the L1 trajectory error does not reflect the observed degradation in model robustness.

\begin{table}
\centering
\begin{tabular}{c|cc|c|c}
\toprule
Augmentation               & DS $\uparrow$      & SR $\uparrow$     & $\text{IR}_s$ $\downarrow$ & L1 (m) $\downarrow$ \\ \midrule
\ding{51}                  & $\mathbf{82.62}$   & $\mathbf{57.43}$  & $\mathbf{0.055}$           & $\mathbf{1.43}$     \\
\ding{55}                  & $66.60$            & $33.64$           & $0.314$                    & $1.47$              \\ \bottomrule
\end{tabular}
\mycaption{Ablation of camera augmentation}{The absence of camera augmentation significantly degrades closed-loop performance, despite minimal impact on open-loop L1 trajectory deviation. This underscores the contribution of our BEV-based augmentation to robust driving and covariate shift mitigation.}
\label{tab:ablation-augmentation}
\end{table}

\myparagraph{BEV Size}
Our tokenizer compresses high-resolution BEV features ($\mathcal{F}_\text{Bev}$) into low-resolution scene tokens ($\mathcal{F}_\text{Scene}$).
An alternative is to directly learn a low-resolution BEV feature space.
As shown in \cref{tab:ablation-bev}, despite exposing the same number of scene tokens to the planner, direct low-resolution BEV generation significantly degrades closed-loop performance.
Specifically, the deformable BEV-image cross-attention of our BEV encoder (based on BEVFormer~\cite{BEVFormerLearningBirds2022}) extracts image information more sparsely when generating low-resolution BEVs, potentially omitting fine details.
Furthermore, it prevents leveraging deformable refinement layers for sampling local, high-resolution features around the future trajectory.
As a result, we observe $7.4 \times$ more static infractions with the low-resolution BEV encoder compared to our compression approach.
This finding underscores the necessity of compressing high-resolution representations rather than directly learning low-resolution ones.

\begin{table}
\centering
\begin{tabular}{cc|cc|c}
\toprule
BEV               & Scene Tokens    & DS $\uparrow$ & SR $\uparrow$ & $\text{IR}_s$ $\downarrow$ \\ \midrule
$100 \times 100$  & $25 \times 15$  & $\mathbf{82.62}$         & $\mathbf{57.43}$         & $\mathbf{0.055}$                      \\
$25 \times 25$    & $25 \times 15$  & $72.18$         & $40.36$         & $0.409$                      \\ \bottomrule
\end{tabular}
\mycaption{BEV Resolution and Tokenization}{High-resolution BEV generation, compressed via our tokenizer, yields superior closed-loop driving performance compared to direct low-resolution BEV generation.}
\label{tab:ablation-bev}
\end{table}

\myparagraph{Number of denoising steps}
The iterative denoising of diffusion models critically impacts the inference latency in real-world deployments.
The runtime of our diffusion-based planner linearly increases with the number of denoising steps $S$, while the point estimator planner has constant runtime, corresponding to $S=1$.
We thus evaluate how $S$ affects closed-loop driving performance and inference FPS in \cref{tab:ablation-denoising}.
In contrast to DiffusionDrive~\cite{DiffusionDriveTruncatedDiffusion2025}, we achieve constant driving performance for $S \in \{2,5,10\}$, without applying their truncated diffusion framework.
We attribute this to a bug in their diffusion schedule, which we detail in the appendix.

\begin{table}
\centering
\begin{tabular}{@{}c|ll|l@{}}
\toprule
$S$ & DS $\uparrow$  & SR $\uparrow$  & FPS $\uparrow$ \\ \midrule
10  & 88.11          & \textbf{72.73} & 4.2            \\
5   & 88.33          & 72.72          & 5.8            \\
2   & \textbf{88.53} & 72.72          & \textbf{7.5}   \\ \bottomrule
\end{tabular}
\mycaption{Impact of Denoising Iterations}{Reducing denoising steps significantly boosts inference FPS while preserving closed-loop driving performance, enabling real-time applications. FPS measured on a Quadro RTX 8000.}
\label{tab:ablation-denoising}
\end{table}

\subsubsection{Multi-Ability Evaluation}

\begin{table*}
\centering
\begin{tabular}{@{}lcccccc@{}}
\toprule
\multirow{2}{*}{\textbf{Method}}                            & \multicolumn{6}{c}{\textbf{Ability} (\%) $\uparrow$}                                \\ \cmidrule(l){2-7} 
                                                            & Merging                           & Overtaking         & Emergency Brake    & Give Way                    & Traffic Sign       & \textbf{Mean}               \\ \midrule
VAD~\cite{VADVectorizedScene2023}                           & 8.11                              & 24.44              & 18.64              & 20.00                       & 19.15              & 18.07                       \\
UniAD-Base~\cite{PlanningorientedAutonomous2023}            & 14.10                             & 17.78              & 21.67              & 10.00                       & 14.21              & 15.55                       \\ \midrule
ThinkTwice~\cite{ThinkTwiceDriving2023}                     & 27.38                             & 18.42              & 35.82              & 50.00                       & 54.23              & 37.17                       \\
DriveAdapter~\cite{DriveAdapterBreakingCoupling2023}        & 28.82                             & 26.38              & 48.76              & 50.00                       & 56.43              & 42.08                       \\
Hydra-NeXt~\cite{HydranextRobust2025}                       & 40.00                             & 64.44              & 61.67              & 50.00                       & 50.00              & 53.22                       \\
Orion~\cite{ORIONHolisticEnd2025}                           & 25.00                             & 71.11              & 78.33              & 30.00                       & 69.15              & 54.72                       \\
TF++~\cite{HiddenBiasesEnd2023}                             & 58.75                             & 57.77              & 83.33              & 40.00                       & 82.11              & 64.39                       \\
Simlingo~\cite{SimLingoVisionOnly2025}                      & 54.01 \small{$\pm 2.63$}          & 57.04 \small{$\pm 3.40$} & \underline{88.33} \small{$\pm 3.34$} & \underline{53.33} \small{$\pm 5.77$}          & \underline{82.45} \small{$\pm 4.73$} & 67.03 \small{$\pm 2.12$}          \\
Hip-AD~\cite{HiPADHierarchical2025}                         & 50.00                             & \textbf{84.44}     & 83.33              & 40.00                       & 72.10              & 65.98                       \\
BridgeDrive$^\dagger$~\cite{BridgeDriveDiffusionBridge2025} & \underline{63.50}                             & 58.89              & \textbf{88.34}     & 50.00                       & \textbf{88.95}     & \underline{69.93}                       \\ \midrule
BevAD-S \textit{(ours)}                                     & 55.83 \small{$\pm 0.72$}          & 53.33 \small{$\pm 6.67$} & 63.33 \small{$\pm 1.93$} & 46.67 \small{$\pm 5.77$}          & 60.88 \small{$\pm 3.08$} & 56.01 \small{$\pm 3.78$}          \\
BevAD-M \textit{(ours)}                                     & \textbf{71.67} \small{$\pm 2.60$} & \underline{74.07} \small{$\pm 1.29$} & 75.56 \small{$\pm 4.41$} & \textbf{76.67} \small{$\pm 5.77$} & 75.44 \small{$\pm 1.61$} & \textbf{74.68 \small{$\pm 1.24$}} \\ \bottomrule
\end{tabular}
\mycaption{Multi-Ability Evaluation}{BevAD consistenly achieves high results across all skills, dominating the mean score score. Notably, it significantly surpasses prior work in the \textit{Merging} and \textit{Give Way} skills. However, it exhibits comparatively lower performance in \textit{Overtaking}, \textit{Emergency Brake}, and \textit{Traffic Sign} skills. \textit{Legend:} $\dagger$: concurrent work.}
\label{tab:multi-ability}
\end{table*}

To gain a nuanced understanding of system performance in closed-loop driving, we employ the fine-granular multi-ability evaluation protocol from Bench2Drive \cite{Bench2DriveTowardsMulti2024}.
This protocol defines five advanced urban driving skills: Merging, Overtaking, Emergency Brake, Give Way, and Traffic Sign. Each of the 220 test routes is mapped to one or more skills necessary for successfully navigating the scenario.
\cref{tab:multi-ability} presents a comparison of BevAD's multi-ability scores against prior work.
BevAD consistently achieves high scores ($>70\%$) across all skills, dominating the mean score.
In contrast, prior state-of-the-art methods \cite{SimLingoVisionOnly2025, HiPADHierarchical2025} exhibit uneven performance, excelling in some skills while underperforming in others.
BevAD-M surpasses the previous best in Merging and Give Way skills by +8.17 and +23.34, respectively.
These skills demand comprehensive surround perception, underscoring BevAD's effective utilization of its multi-view camera system.
However, BevAD lacks in terms of Overtaking, Emergency Brake, and Traffic Sign skills compared to the best prior or concurrent methods for each skill.

\begin{figure}
    \centering
    \includegraphics[width=\linewidth]{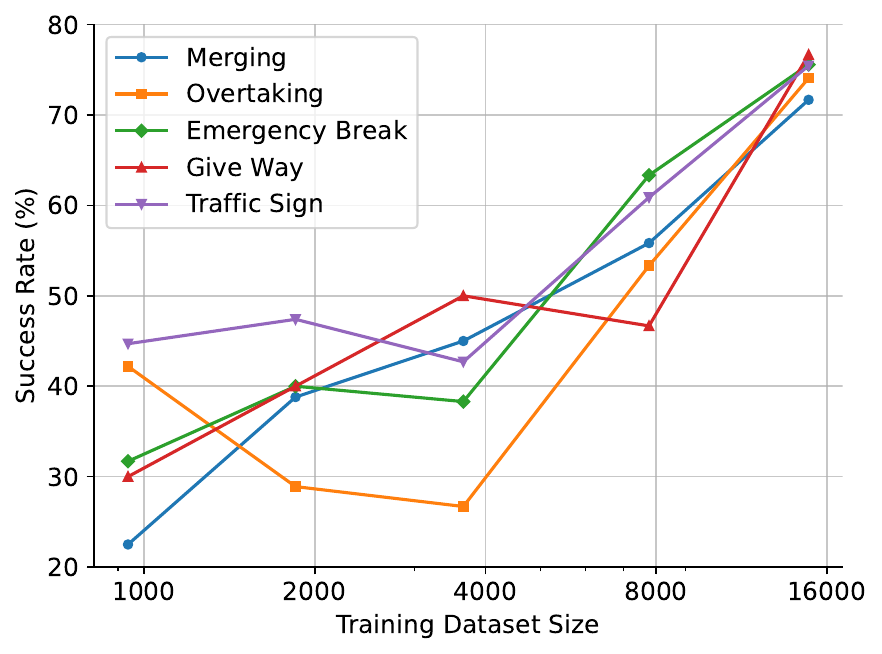}
    \mycaption{Emerging Skills}{All skills remain underdeveloped with fewer than 4,000 training episodes. As the volume of training data increases, the skills progressively and uniformly emerge.}
    \label{fig:scaling-skills}
\end{figure}

Building on the discussion in \cref{sec:scalability}, we present the progression of BevAD's closed-loop driving skills as the training dataset size increases, detailed in \cref{fig:scaling-skills}.
With fewer than 4,000 training episodes, all skills exhibit a success rate below 50\%.
However, as the training data is doubled and quadrupled, the skills show consistent and uniform improvement.
This enhancement is evidenced by high success rates in complex scenarios, such as overtaking amidst oncoming traffic, merging onto highways and into traffic flow at intersections, and yielding to emergency vehicles.

\subsubsection{Qualitative Results}

\begin{figure*}
    \centering
    \includegraphics[width=1\linewidth]{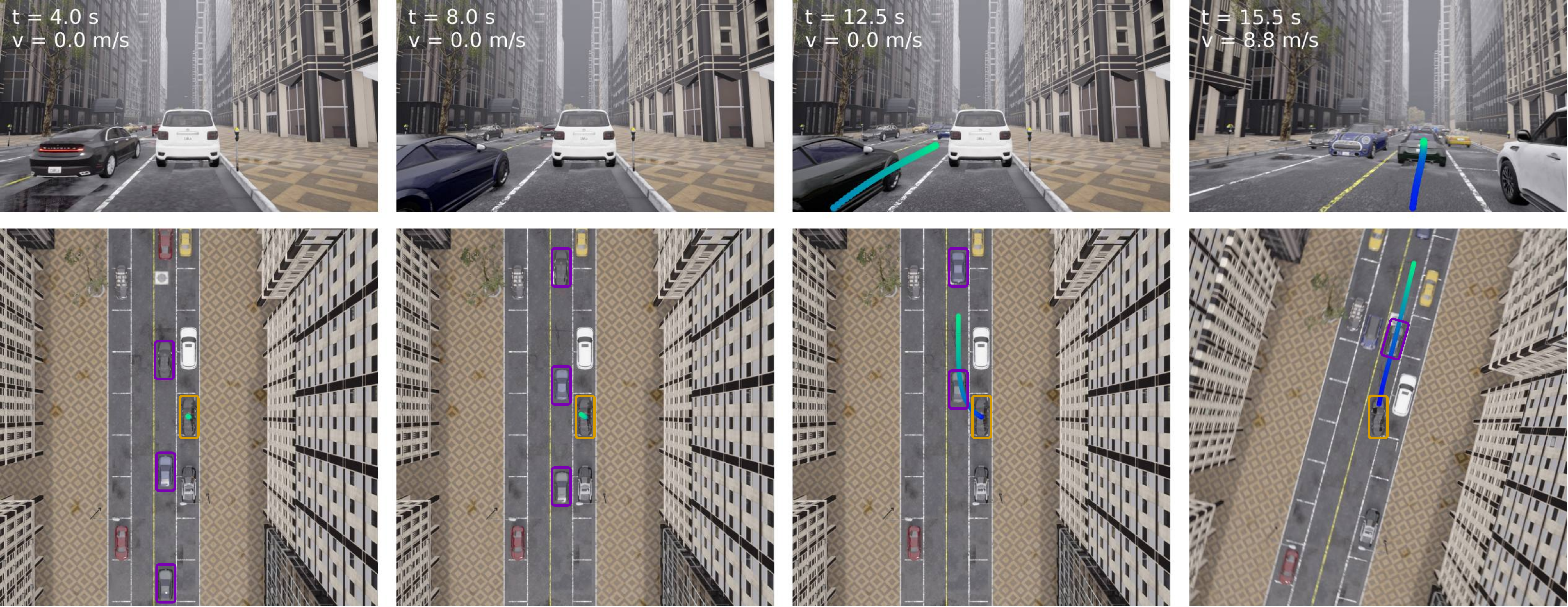}
    \mycaption{Merging}{BevAD merges from a parallel parking space into traffic. It yields to rear-end flow of vehicles, identifies a safe gap, and accelerates for seamless merging.}
    \label{fig:demo-parkingexit}
\end{figure*}

\begin{figure*}
    \centering
    \includegraphics[width=1\linewidth]{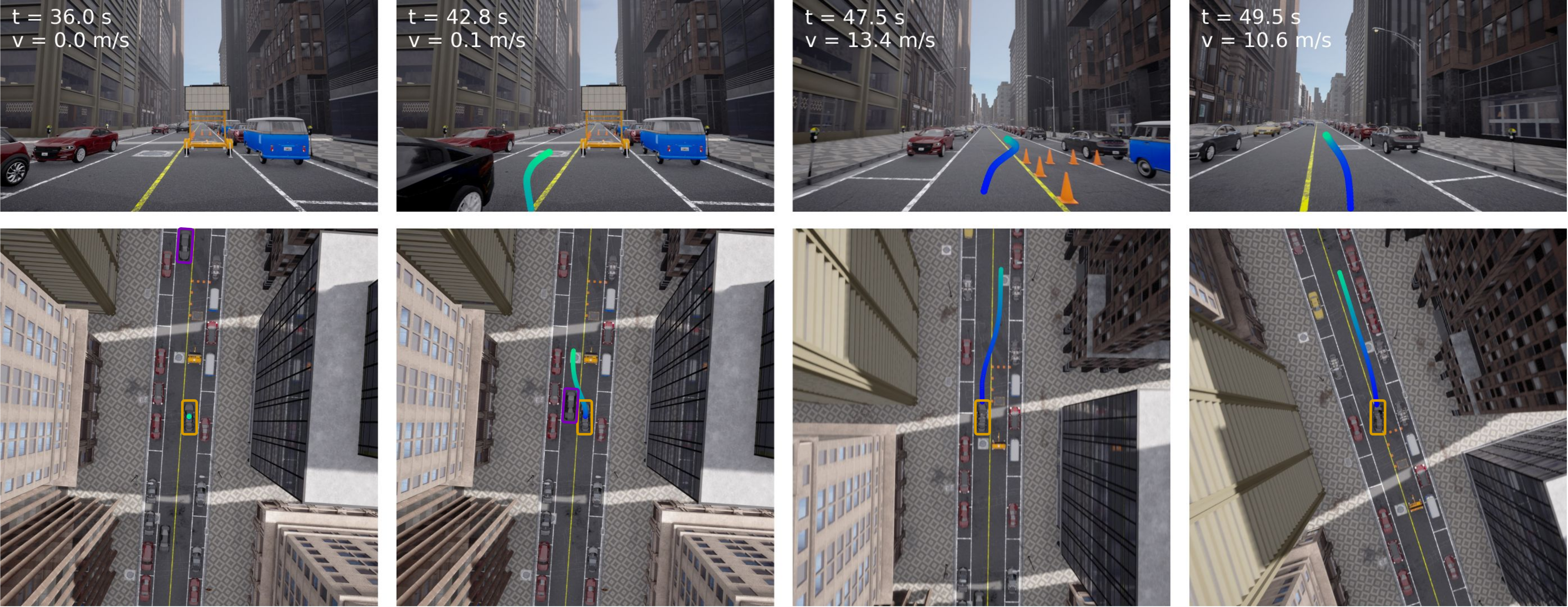}
    \mycaption{Overtaking (1)}{BevAD executes an overtaking maneuver when the route is blocked by a construction vehicle. It waits for clear oncoming traffic, then steers into the opposing lane, accelerates to quickly pass the obstacle, and subsequently decelerates when returning to its original lane.}
    \label{fig:demo-construction}
\end{figure*}

\begin{figure*}
    \centering
    \includegraphics[width=1\linewidth]{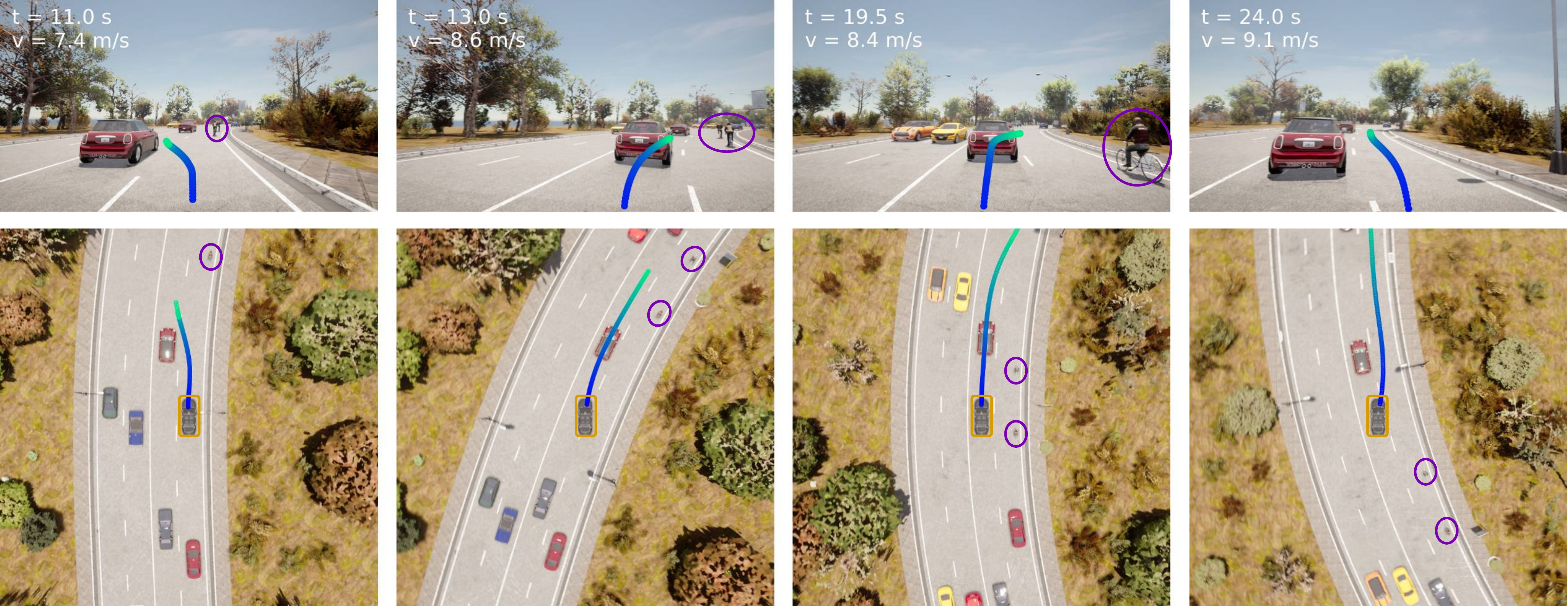}
    \mycaption{Overtaking (2)}{BevAD approaches a group of cyclists. It executes a safe left lane change to overtake them. After the maneuver, BevAD returns to its original lane, maintaining a safe distance to the cyclists.} \label{fig:demo-hazard}
\end{figure*}

\begin{figure*}
    \centering
    \includegraphics[width=1\linewidth]{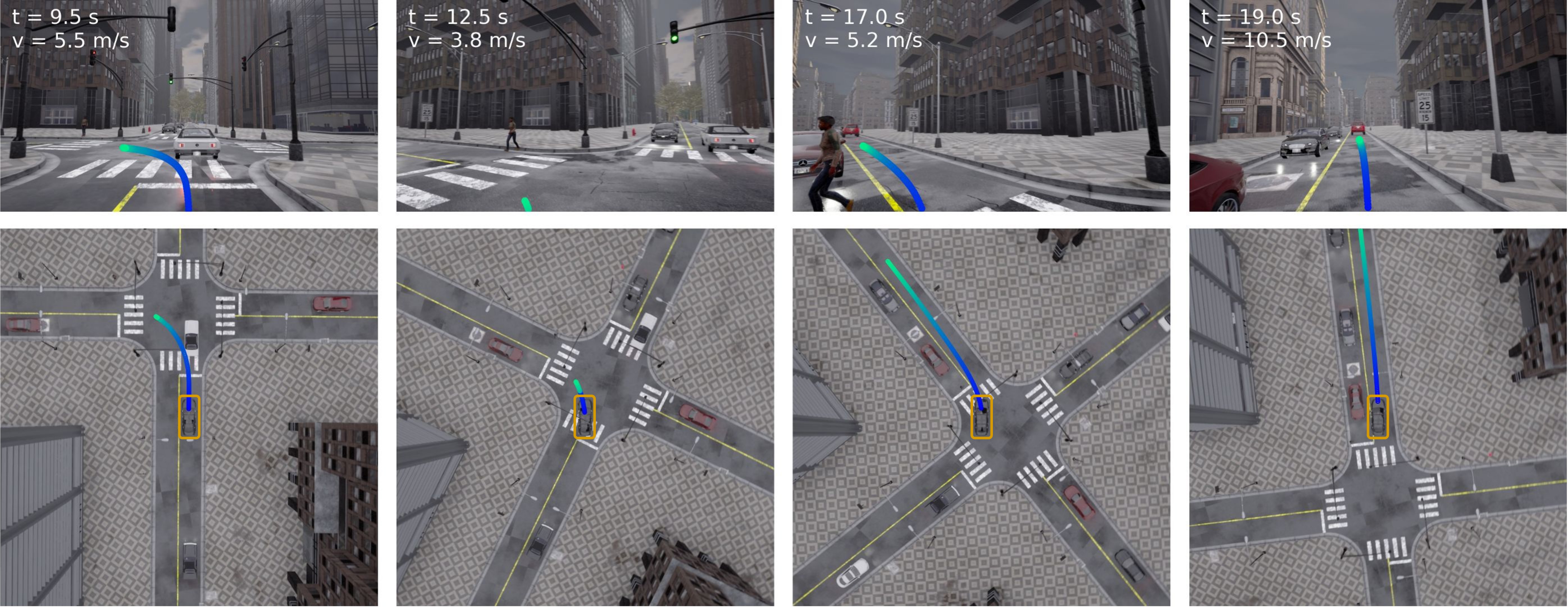}
    \mycaption{Emergency Brake}{BevAD brakes at a green-light intersection due to a pedestrian crossing its left-turn path. Driving resumes upon pedestrian clearance.}
    \label{fig:demo-pedestrian}
\end{figure*}

\begin{figure*}
    \centering
    \includegraphics[width=1\linewidth]{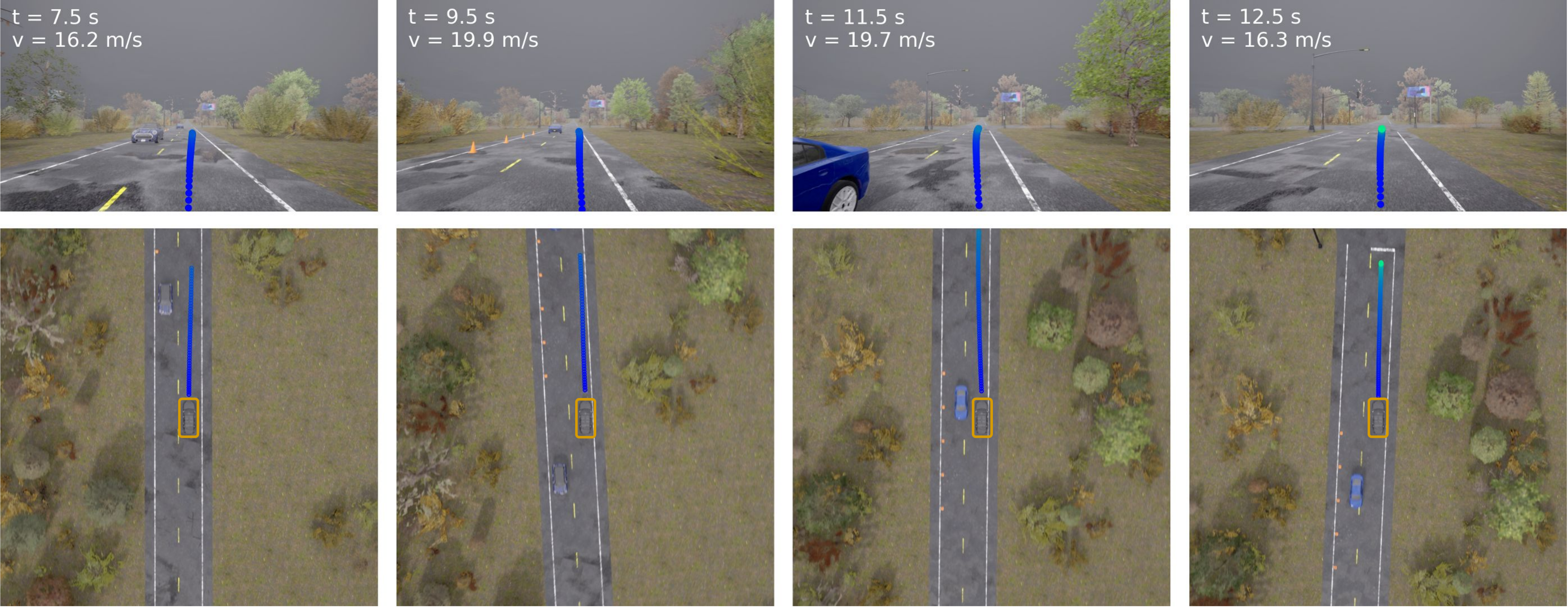}
    \mycaption{Give Way}{BevAD yields to an oncoming vehicle encroaching on the ego lane. It performs a controlled rightward deviation from the lane center, staying within road limits, and re-centers once the oncoming traffic has passed.}
    \label{fig:demo-invadingturn}
\end{figure*}

\begin{figure*}
    \centering
    \includegraphics[width=1\linewidth]{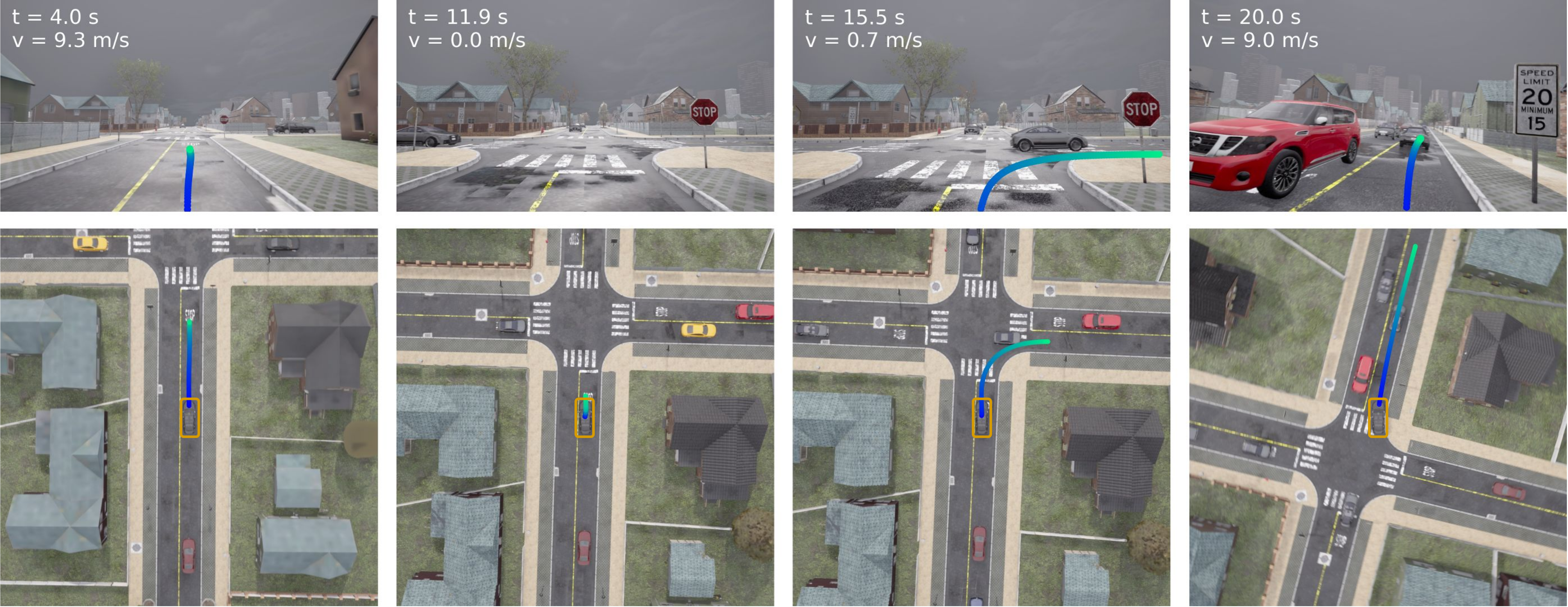}
    \mycaption{Traffic Sign}{BevAD stops at a stop sign at an intersection with cross-traffic. It waits for a safe gap, then executes a right turn and merges into the traffic flow.}
    \label{fig:demo-stopsign}
\end{figure*}

We provide qualitative closed-loop driving examples for each multi-ability skill of BevAD-M in \cref{fig:demo-parkingexit}, \ref{fig:demo-construction}, \ref{fig:demo-hazard}, \ref{fig:demo-pedestrian}, \ref{fig:demo-invadingturn} and \ref{fig:demo-stopsign}.
The examples are best viewed when zoomed in and visualize planned ego trajectories that are generated by rolling out the predicted speed profile along the predicted path.
This depiction aids in understanding the dynamic planning component, but is solely for visualization.
It does not serve as controller input, nor does it affect the lateral accuracy of the predicted path.

\onecolumn

\twocolumn

\end{document}